\documentclass[journal]{IEEEtran}
\usepackage{times}
\usepackage{url}
\usepackage{epsfig}
\usepackage{caption}
\usepackage{graphicx}
\usepackage{amsmath}
\usepackage{amssymb}
\usepackage{subfigure}
\usepackage{multirow}
\usepackage{color}
\usepackage{amsmath}
\usepackage{subfigure}
\usepackage{algorithm}
\usepackage{algorithmic}
\usepackage{relsize}
\usepackage{afterpage}
\ifCLASSINFOpdf
\else
\fi
\begin{document}
\title{High Quality Image-based Face Inpainting using Domain Embedded Multi-model Generative Adversarial Networks}
%Xian

\author{
Xian Zhang$^{*}$,
Xin Wang$^{*}$,~\IEEEmembership{Member,~IEEE},
Bin Kong,
Canghong Shi,
Youbing Yin,\\
Qi Song,
Siwei Lyu,~\IEEEmembership{Senior Member,~IEEE},
Jiancheng Lv,~\IEEEmembership{Member,~IEEE},
Xi Wu,
Xiaojie Li$^{\dagger}$,~\IEEEmembership{Member,~IEEE}
\thanks{X. Zhang, X. Wu, X. Li are with the Chengdu University of Information Technology, China. X. Wang, B. Kong, Y. Yin and Q. Song are with the CuraCloud Corporation, Seattle, USA. S. Lyu is with the SUNY Albany, NY, USA. C. Shi is with the Southwest Jiaotong University, China. J. Lv is with the Sichuan University, China. }
\thanks{$^{*}$ Xian Zhang and Xin Wang are co-first authors.}
\thanks{$^{\dagger}$ Corresponding authors: lixj@cuit.edu.cn.}
}


\markboth{Journal of \LaTeX\ Class Files,~Vol.~14, No.~8, August~2015}%
{Shell \MakeLowercase{\textit{et al.}}: Bare Demo of IEEEtran.cls for IEEE Journals}



\maketitle

\begin{abstract}
Prior knowledge of face shape and structure plays an important role in face inpainting.
However, traditional face inpainting methods mainly focus on the generated image resolution of the missing portion without consideration of the special particularities of the human face explicitly and generally produce discordant facial parts. To solve this problem, we present a domain embedded multi-model generative adversarial model for inpainting of face images with large cropped regions.
We firstly represent only face regions using the latent variable as the domain knowledge and combine it with the non-face parts textures to generate high-quality face images with plausible contents. Two adversarial discriminators are finally used to judge whether the generated distribution is close to the real distribution or not. It can not only synthesize novel image structures but also explicitly utilize the embedded face domain knowledge to generate better predictions with consistency on structures and appearance. Experiments on both CelebA and CelebA-HQ face datasets demonstrate that our proposed approach achieved state-of-the-art performance and generates higher quality inpainting results than existing ones.
\end{abstract}

\begin{IEEEkeywords}
Face Inpainting, Domain Embedding, High Quality Image, Adversarial Generative Model.
\end{IEEEkeywords}

% \fi
\IEEEpeerreviewmaketitle

\section{Introduction}

\begin{figure}[!t]
    %\vskip 0.2in
    \begin{center}

        \begin{minipage}[t]{0.15\textwidth}
            \includegraphics[width=\linewidth]{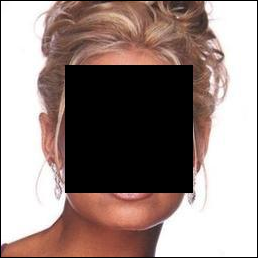}
                                    \centerline{(a) Cropped}
        \end{minipage}
        \begin{minipage}[t]{0.15\textwidth}
            \includegraphics[width=\linewidth]{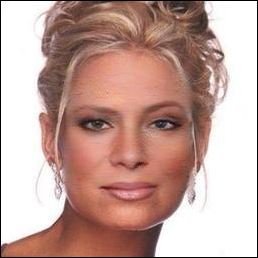}
                                    \centerline{(b) Ours}
        \end{minipage}

        \begin{minipage}[t]{0.15\textwidth}
            \includegraphics[width=\linewidth]{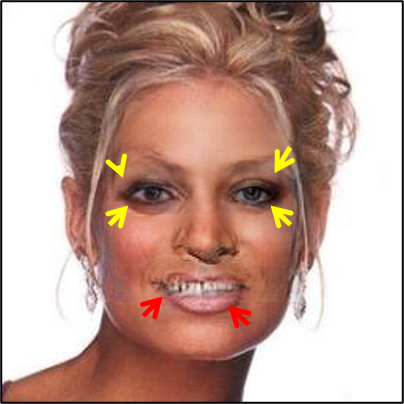}
                                    \centerline{(c) CE~\cite{pathak2016context}}
        \end{minipage}
         \begin{minipage}[t]{0.15\textwidth}
            \includegraphics[width=\linewidth]{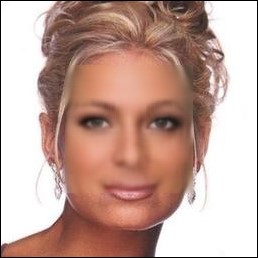}
                                    \centerline{(d)  GLCIC~\cite{iizuka2017globally}}
        \end{minipage}
        \begin{minipage}[t]{0.15\textwidth}
            \includegraphics[width=\linewidth]{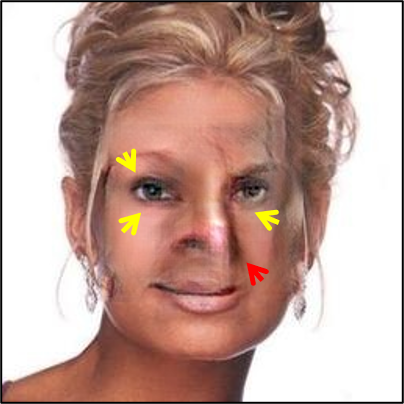}
                                    \centerline{(e)  CA~\cite{yu2018generative}}
        \end{minipage}

        \begin{minipage}[t]{0.15\textwidth}
            \centerline{\includegraphics[width=\linewidth]{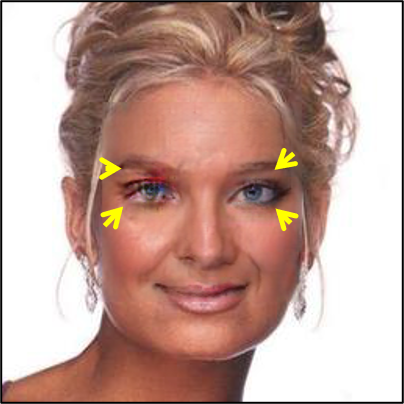}}
            \centerline{(f) PICNet~\cite{zheng2019pluralistic}}
        \end{minipage}
        \begin{minipage}[t]{0.15\textwidth}
            \centerline{\includegraphics[width=\linewidth]{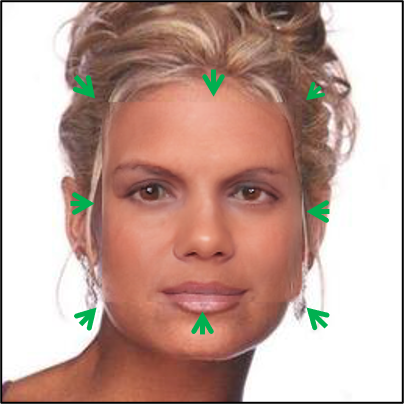}}
            \centerline{(g) PEN~\cite{yan2019PENnet}}
        \end{minipage}
        \begin{minipage}[t]{0.15\textwidth}
            \centerline{\includegraphics[width=\linewidth]{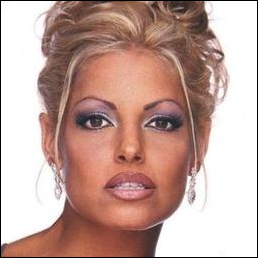}}
            \centerline{(h) Original}
        \end{minipage}
    \end{center}
    \caption{Challenging example handled by previous approaches.
   (a) cropped image (h) original image. In result (c), the facial part has fuzzy boundaries and details (red arrow), which is not visually-realistic. In result (d), the completion area is too blurry. In results (e) and (f), some parts are not structurally-reasonable, e.g. the eyes (yellow arrow) are asymmetry with different size. In the result (g), the completion has obvious boundary artifacts (green arrow). Our result (b) shows excellent performance on texture and facial structure.}
\label{intro_qualitative}
\vskip -0.2in
\end{figure}
\IEEEPARstart{I}{mage} inpainting refers to the task of filling in missing or masked regions with synthesized contents.
%{\color{red}Traditional machine learning methods shrink the missing area by the weighted sum of the surrounding pixels until the missing area is completed or copy and realign the background patches to complete the holes. PatchMatch (PM)~\cite{barnes2009patchmatch} is to realize image completion in this way. But this method is difficult to complete when there are large missing blocks (see Fig.~\ref{intro_qualitative}(c)).}
Among the various ways of vision algorithm of today, deep learning
based methods have attracted a lot of attention in image inpainting.
The earliest deep learning image inpainting method was called context encoders (CE) by Deepak Pathak et al.~\cite{pathak2016context}. They compulsively obtain latent characteristic information of the missing area by context information.
However, the context encoders only pay attention to the missing area information rather than the whole image such that the generated image would have obvious patching marks at the boundary (see Fig.~\ref{intro_qualitative}(c)). To overcome this limitation, Iizuka et al. proposed globally and locally consistent image completion (GLCIC)~\cite{iizuka2017globally}. GLCIC uses the missing region discriminator and the global discriminator to ensure the consistency of the spatial distribution of the global and the local, but it can only complete the regular region for inpainting. Yu et al.~\cite{yu2018generative} proposed generative image inpainting with contextual attention (CA), it first generated a low-resolution image in the missing area, then updated the refinement image by searching for patches similar to an unknown area from a known area with contextual attention.
Zheng et al.~\cite{zheng2019pluralistic} proposed a pluralistic image completion network (PICNet) with a reconstructive path and the generative path to creating multiple plausible results.  Zeng et al.~\cite{yan2019PENnet} proposed a pyramid-context encoder network (PEN) with pyramid structure to complete image.
However, all these methods produce discordant facial parts, which are not structurally-reasonable. For example, the asymmetry eyebrow (see Fig.~\ref{intro_qualitative}(c)), blurry result (see Fig.~\ref{intro_qualitative}(d)), one eye is large and the other one is small (see Fig.~\ref{intro_qualitative}(e)), two eyes of one people have different colors (see Fig.~\ref{intro_qualitative}(f)) or boundary artifacts (see Fig.~\ref{intro_qualitative}(g)).

One possible reason is that these general image inpainting methods mainly focus on the resolution of the generated image but without consideration of the special particularities of the human face (e.g., symmetrical relation, harmonious relation) in their approach.
{Another reason is that only a few works~\cite{li2017generative,liao2018face} dedicated to the task of face inpainting.} The current face inpainting algorithms incorporate simple face features into a generator for human face completion. However, the benefit of face region domain information has not been fully explored, which also leads to unnatural images.
Face inpainting remains a challenging problem as it requires to generate semantically new pixels for the missing key components with consistency on structures and appearance.

In this paper, A Domain Embedded Multi-model Generative Adversarial Network (DEGNet) is proposed for face inpainting.
First, we embed the face region domain information (i.e., face mask, face part and landmark image) by variational auto-encoder into a latent variable space as the guidance, where only face features lie in the latent variable space.
 After that, we combine the face region domain embedded latent variable into the generator for face inpainting. Finally, a global discriminator and patch discriminator~\cite{isola2017image} are finally used to judge whether the generated distribution is close to the real distribution or not. Experiments on two benchmark face datasets~\cite{liu2015faceattributes,karras2018progressive} demonstrate that our proposed approach generates higher quality inpainting results than the state-of-the-art methods. The main contributions of this paper are summarized as follows:

\begin{itemize}

  \item  Our proposed model embeds the face region information into latent variables as the guidance information for face inpainting. The proposed method produces sharper and more natural faces, and thus leads to improved face structure prediction, especially for larger missing regions.

  \item We design a new learning scheme with the patch and global adversarial loss, in which the global discriminator could control the overall spatial consistency and the patch discriminator~\cite{isola2017image} could provide a  more elaborate face feature distribution which can generate impressively photorealistic high-quality face images.

  \item To the best of our knowledge, our work is the first on the evaluation of the side face inpainting problem, and more importantly, our impainting results of side face show excellent visual quality and facial structures comparing to the state-of-the-art methods.

\end{itemize}

\begin{figure*}[t]
    \setlength{\abovecaptionskip}{-0.02cm}
    \centering
    \includegraphics[width=0.98\textwidth]{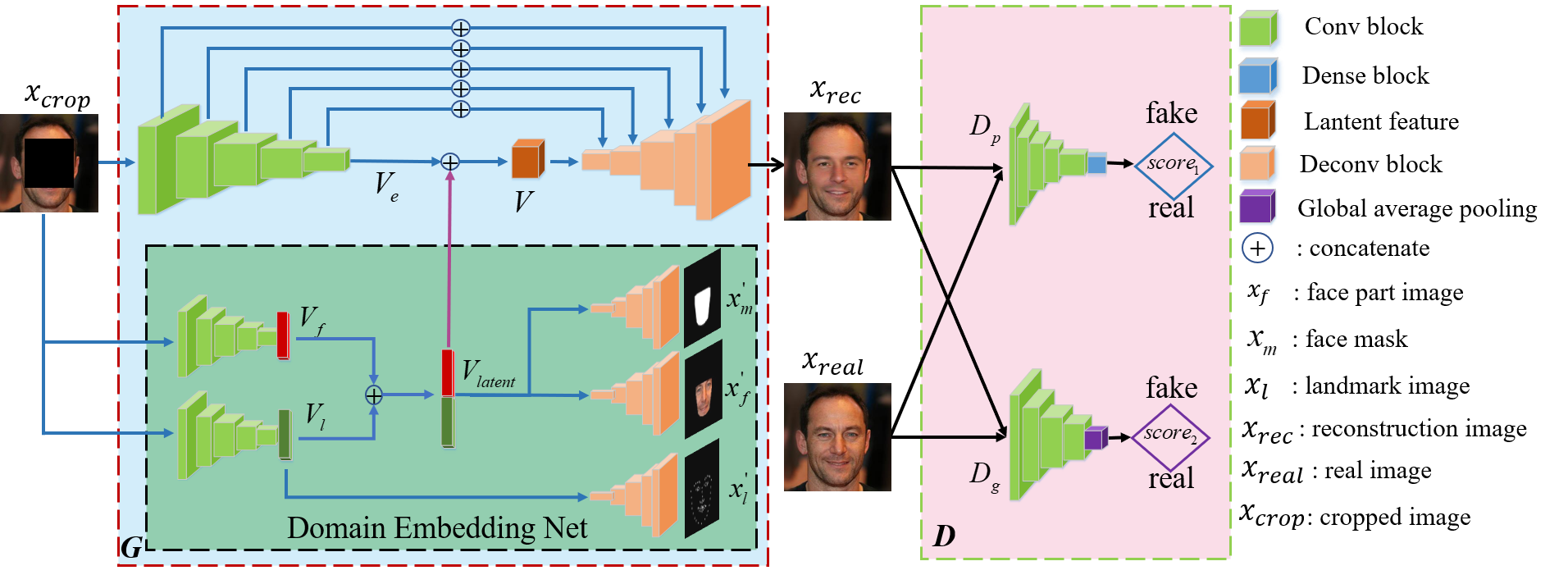}
    \caption{Overview of the DEGNet architecture, it mainly consists of three modules: Domain Embedding Net, Domain Embedded Generator (G) and Multi-model Discriminators (D), \textit{i.e.} $D_g$ (global discriminator) and $D_p$ (patch discriminator).}
    \label{fig_network}
\end{figure*}

\section{Related Work}
\textbf{General Image Inpainting.} Traditional diffusion-based or patch-based methods~\cite{bertalmio2000image,ballester2001filling,bertalmio2003simultaneous,jin2015annihilating,dobrosotskaya2008wavelet,qin2013novel,xu2010image,criminisi2004region,li2014color,liu2012exemplar,li2014universal,ballester2007inpainting} with low-level features generally assume image holes share similar content to visible regions; thus they would directly match, copy and realign the background patches to complete the holes.
These methods perform well for background completion,
e.g. for object removal, but cannot hallucinate unique
content not present in the input images.
Barnes et al.~\cite{barnes2009patchmatch} proposed a fast nearest-neighbor field algorithm called
PatchMatch (PM) for image editing applications including inpainting.
It greatly reduces the search scopes of patch similarity by utilizing
the continuity of images.
Based on the nearest neighbor (NN) inpainting method,
Whyte et al.~\cite{whyte2009get} updated the predicted region by finding the nearest image
from the original image in the training dataset.
While the above methods become less effective
when the missing region becomes large or irregular.

%-------------------------------------------------------------------------
Many recent image inpainting methods are proposed based on deep learning model~\cite{jin2015annihilating,he2016deep,yang2017high, liu2018image,yan2018shift,ren2019structureflow,hong2019deep,xie2019image}. Li et al.~\cite{li2017generative} propose a deep generative network for face completion, it consists of an encoding-decoding generator and two adversarial discriminators to synthesize the missing contents from random noise.
The proposed model~\cite{xiong2019foreground,song2018contextual,yan2018shift,wang2018image,yu2019free,ding2018image,xue2017depth,buyssens2016depth} can synthesize plausible contents for
the missing facial key parts from random noise. Iizuka et al.~\cite{iizuka2017globally} proposed GLCIC that combines a global discriminator with a missing region discriminator to ensure that the missing area is consistent with the whole image and it used dilated convolution to increase the receptive field.
Alternatively, given a trained generative model, Raymond et al.~\cite{yeh2017semantic}  Search for the closest encoding of the corrupted image in the latent image manifold using their context and prior losses to infer the missing content by the generative
model.
To recover large missing areas of an image, Patricia et al.~\cite{vitoria2018semantic}
tackle the problem not only by using the available visual data but also by taking advantage
of generative adversarial networks incorporating image semantics.
However, these methods can generate visually plausible image structures and textures, but usually, create distorted structures or blurry textures inconsistent with surrounding areas.
% {\color{red} Iizuka et al.  \cite{iizuka2017globally} came up with a method called GLCIC combines the local with the whole to make a generative antagonism to ensure that the missing area is consistent with the spatial distribution of the whole image. However, the completed image by GLCIC is too blurry. Iizuka S et al. proposed using Dilated convolution to increase the receptive field. To create more realistic and clear content in the cropped field, they used global discriminator and local discriminator to optimize the whole image and cropped field respectively.}

To reduce the blurriness issue commonly existing in the CNN-based inpainting, two-stage methods have been proposed to conduct texture refinement on the initially completed images~\cite{zhang2018semantic,yu2018generative,song2018contextual}. %by re-placing the neural patch in the predicted region using the closest one in the known region.
Generally, they firstly filled the missing regions by a content generation network and then updated the neural patch in the predicted region with fine textures in the known region.
Recently Yu et al.~\cite{yu2018generative} propose a new deep generative model-based approach for inpainting. It not only synthesize novel image structures but also explicitly utilize surrounding image features as references to make better predictions. While it is likely to fail when the source image does not contain a sufficient amount of data to fill in the unknown regions. When the training image is a Non-HQ image, it performs not well.
Furthermore, such processing might introduce undesired content change in the predicted region, especially when the desired content does not exist in the known region.
To avoid generating such in-correct content, Xiao et al.~\cite{xiao2019cisi} propose a content inference and style imitation network for image inpainting. It explicitly separates the image data into content code and style code to generate the complete image.
It performs well on structural and natural images in terms of content accuracy as well as texture details but does not demonstrate its performance on face image inpainting.
%FSNet \cite{Natsume2018FSNet}
Zheng et al.~\cite{zheng2019pluralistic} present a pluralistic image completion approach for generating multiple and diverse plausible solutions for image completion.
However, it cannot keep stable performance and need a sufficiently varied high-quality dataset.
%{\color{red}Zeng et al.~\cite{yan2019PENnet} think that the information in the previous features is not fully utilized, so a pyramid-context encoder network is used to complete image inpainting. But there are obvious border marks.}

\textbf{Face Inpainting.} Li et al. proposed a deep generative face inpainting model that consists of an encoding-decoding generator, two adversarial discriminators, and a parsing network to synthesize the missing contents from random noise. To ensure pixel faithfulness and local-global contents consistency, they use an additional semantic parsing network to regularize the generative networks (GFCNet)~\cite{li2017generative}.
In 2018, under a novel generative framework called collaborative GAN (collagen)~\cite{liao2018face}, Liao et al. proposed a collaborative adversarial learning approach to facilitate the direct learning of multiple tasks including face completion, landmark detection and semantic segmentation for better semantic understanding to ultimately yield better face inpainting.
%{\color{red} Jo Y et al. proposed SC-FEGAN \cite{jo2019sc} to complete the cropped image by using sketch domains and color domains in the cropped field as guided and using Gated convolution to the cropped field.}

\section{Domain Embedded Multi-model GAN}
An overview of our proposed framework is shown in Fig.~\ref{fig_network}. Our goal is to use the domain information of the face part to generate high-quality face inpainting images. We first introduce our Face Domain Embedded Generator and Multi-model Discriminator in Section~\ref{faceG} and Section~\ref{faceD}, respectively. Then, the loss functions of these components are described in Section~\ref{faceLoss}.
Note that the details of network structure (number of layers, size of feature maps, etc.) can be found in the supplementary document (see TABLE~\ref{tab1}).

\subsection{Face Domain Embedded Generator}
\label{faceG}
\begin{figure}[t]
    %\vskip 0.2in
    \begin{center}
        \begin{minipage}[t]{0.07\textwidth}
            \includegraphics[width=\linewidth]{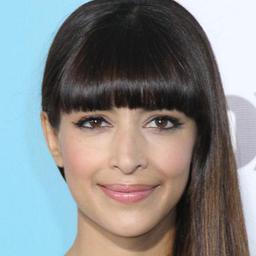}
                                    \centerline{(a)}
        \end{minipage}
        \begin{minipage}[t]{0.07\textwidth}
            \includegraphics[width=\linewidth]{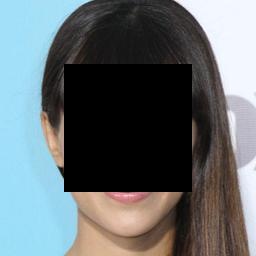}
                                    \centerline{(b)}
        \end{minipage}
        \begin{minipage}[t]{0.07\textwidth}
            \includegraphics[width=\linewidth]{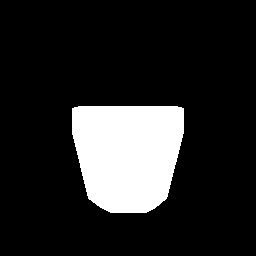}
                                    \centerline{(c)}
        \end{minipage}
        \begin{minipage}[t]{0.07\textwidth}
            \includegraphics[width=\linewidth]{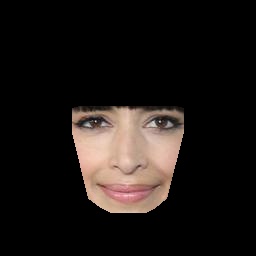}
                                    \centerline{(d)}
        \end{minipage}
        \begin{minipage}[t]{0.07\textwidth}
            \centerline{\includegraphics[width=\linewidth]{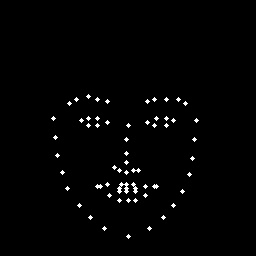}}
            \centerline{(e)}
        \end{minipage}
        \begin{minipage}[t]{0.07\textwidth}
            \centerline{\includegraphics[width=\linewidth]{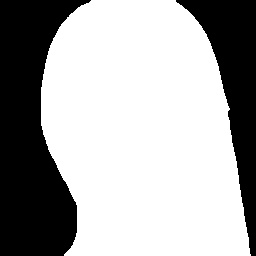}}
            \centerline{(f)}
        \end{minipage}
    \end{center}
    \caption{Our method uses six types of images, (a) Original image, (b) Cropped image,  (c) Face mask image, (d) Face part image, (e)
     Face landmark image, (f) Face foreground mask image. }
    \label{fig_dataset}
    \vskip -0.1in
\end{figure}

\noindent \textbf{Domain Embedding Network.} Given a face image $x_{real}$ and its cropped image $x_{crop}$, our goal is to learn
the most important face domain information to guide the inpainting procedure.
We use face region images which include face mask image $x_m$ and face part image $x_f$  and face landmarks image $x_l$ to represent the face domain information, see Fig. \ref{fig_dataset} (c,d,e).
The reason we use face masks, face part and landmarks as face domain information is that the outline of face information can be extracted from the mask, face semantic information can be extracted from the face part and face structure information can be extracted from the face landmarks.

Then, we use a VAE network~\cite{doersch2016tutorial,pu2016variational,lutz2018alphagan} with an encoder-decoder architecture to embed the face domain information into latent variables.
%Its detailed structure is illustrated in Figure \ref{fig_network} (b).
More specifically, in the encoding phase, the corrupted face image $x_{crop}$ is first passed to two encoders, face region encoder and face landmark encoder, yielding standard normal distributions for the face region image and face landmarks image, respectively. Subsequently, latent variables are sampled from each of these two normal distributions:
\begin{equation}
\label{eqzconstri}
\begin{aligned}
    &V_{f}\sim\mathcal{N}(\mu_{f},\sigma_{f})\\
    &V_{l}\sim\mathcal{N}(\mu_{l},\sigma_{l})
\end{aligned}
\end{equation}
%where $\mu_{\theta-*}$ and $\sigma_{\theta-*}^2$ are the mean and variance.
where $V_{f}$ and $V_{l}$ are the sampled latent variables for the face region (Fig. \ref{fig_dataset} (c)), face mask and face landmarks (Fig. \ref{fig_dataset} (d,e)), respectively. $\mu_{\alpha}$ and $\sigma_{\alpha}$ with $\alpha \in\{f,l\}$ denote the means and variances of the generated standard normal distributions. In the decoding phase, $V_{f}$ and $V_{l}$ are concatenated to $V_{latent}$.
The face mask decoder and face part decoder takes $V_{latent}$ as input and generate the reconstructed face mask $x_{m}'$, face part image $x_{f}'$, and the face landmark decoder take $V_{l}$ as input and generate the reconstructed face landmark image $x_{l}'$
(See Fig.~\ref{fig_network}).

During the training process, the above crucial face information is embedded into the latent variable $V_{latent}$. The structure of these two encoders is symmetrical to that of the three decoders and all of them have different weights.  The main reason we employ two encoders and three decoders rather than three encoders and three decoders in domain embedding network are that we don't want the structural information, semantic information, and outline information of the face to be embedded independently, we expect both $V_{l}$ and $V_{f}$ to contain all three kinds of information.

\textbf{Domain Embedded Generator.}
To reconstruct the complete and harmonious face image, we need to integrate the embedded latent variable $V_{latent}$ into the face Generator. We use FCN as our generator, in the encoding phase, the cropped image $x_{crop}$ is sent into FCN and get a latent feature $V_{e}$ with size (16, 16, 512) from the middle layer of the FCN. To concatenate $V_{latent}$ and $V_{e}$, we resize $V_{latent}$ into (16, 16, 2) and concatenate $V_{e}$ and $V_{latent}$ on their last channels, denote as $V$. In the decoding phase, we generate realistic face images by deconvolution blocks with $V$ as input, See Fig.~\ref{fig_network}.
%\IEEEpubidadjcol

\subsection{Multi-model Discriminator}
\label{faceD}
Our DEGNet has two discriminators i.e. global discriminator and patch discriminator from PatchGAN~\cite{isola2017image}. Different from the missing region discriminator in~\cite{iizuka2017globally}, which only focuses on the cropped region. The patch discriminator in our model cuts the entire image into small patches and critics to each patch is real or generated. Thus, the global discriminator forces the generator to get a clear face image, and the patch discriminator to enhance the visual quality of local details of the generated face image.

In particular, when generating face image $x_{rec}$, $x_{rec}$ and $x_{real}$ are passed into two discriminators to distinguish the generated face image is true or fake. The global discriminator can guarantee the spatial consistency of the global structure of $x_{rec}$ with $x_{real}$ in the beginning process. When $x_{rec}$ with $x_{real}$ has been consistent with the overall spatial structure, patch discriminator then split $x_{rec}$ into patches to refine the spatial structure consistent with $x_{real}$ on every patch.

% {\color{red} Its performance is not great When using only global discriminator. It is inefficient when using only patch discriminator. Therefore, the multi-model discriminator is employed to ensure quality and efficiency.}

% {\color{red} While generating realistic face images, the Domain Embedding Net enhances the robustness and generalization performance of the generator and achieves more accurate processing of face information in the missing region under the influence of two discriminators.}

\subsection{Loss Function}
\label{faceLoss}
\noindent \textbf{Domain Embedding Network Loss.} For corrupted face image $x_{crop}$, the VAE network are trained to reconstruct the face mask $x_{m}'$, face region $x_{f}'$, and the landmark image $x_{l}'$. In this work, we define three reconstruction losses (see Eq.~(\ref{eqone})) for these three outputs, respectively:

\begin{equation}
\label{eqone}
\begin{aligned}
    &\mathcal{L}_{m}^{rec}=\mathop{\mathbb{E}}[\mathcal{L}_{CE}(x_m,x_m')]\\
    &\mathcal{L}_{f}^{rec}=\mathop{\mathbb{E}}[||x_f-x_f'||_{1}]\\
    &\mathcal{L}_l^{rec}=\mathop{\mathbb{E}}[\mathcal{L}_{CE}(x_l,x_l')]
\end{aligned}
\end{equation}
where $\mathcal{L}_{CE}$ denotes the cross-entropy loss, $||\cdot||_1$ is the $L_1$ loss, and $x_m$, $x_f$, and $x_l$ are the corresponding ground truth images. The loss functions $\mathcal{L}_{m}^{rec}$, $\mathcal{L}_{f}^{rec}$ and $\mathcal{L}_l^{rec}$ in Eq.~(\ref{eqone}) enforce the domain embedding net learns the missing face profile information, semantic information, and structural information, respectively.

%The encoder can extract face domain information more accurately under the constraint of (\ref{eqone}).

To impose a domain distribution (in our case, the standard normal distribution) on the latent space, we employ a latent classifier $C_\omega$ rather than the Kullback-Leibler divergence used in standard VAEs. This technique has been demonstrated to help the VAE to capture better data manifold, thereby learning better latent representations~\cite{makhzani2015adversarial}, which has also been widely used in various VAE-based generative networks such as $\alpha$-GAN~\cite{rosca2017variational}. The latent classification loss is defined as follows:

\begin{equation}
\label{eqthree}
\begin{aligned} &\mathcal{L}_f^{lat}=-\mathop{\mathbb{E}}[\log C_\omega(V)]-\mathop{\mathbb{E}}[\log(1-C_\omega(V_f))]\\
&\mathcal{L}_t^{lat}=-\mathop{\mathbb{E}}[\log C_\omega(V)]-\mathop{\mathbb{E}}[\log(1-C_\omega(V_l))]\\
&\mathcal{L}_m^{lat}=-\mathop{\mathbb{E}}[\log C_\omega(V)]-\mathop{\mathbb{E}}[\log(1-C_\omega(V_{latent}))]\\
\end{aligned}
\end{equation}
where $V\sim \mathcal{N}(0,1)$ is a random variable randomly sampled from the standard normal distribution. $V_{l}$ and $V_{f}$ are obtained by sampling from two encoders respectively. $V_{latent}$ is connected by $V_{l}$ and $V_{f}$. The function of the Eq.~(\ref{eqthree}) is to enforce $V_{l}$, $V_{f}$ and $V_{latent}$ close to the normal distribution.

% Based on $\alpha$-GAN's~\cite{rosca2017variational}, we employed a discriminator instead of KL divergence to make it adversarial with encoder-decoder, so as to enhance the robustness of the domain embedding network.

\noindent \textbf{Domain Embedded Generator Loss.} To enhance the foreground region that including face and hair regions (Fig.~\ref{fig_dataset}(f)), we add a weight term of foreground and background to the standard reconstruction loss. The weight $W_{FB}$ is defined as follows:
\begin{equation}
\begin{aligned}
 W_{FB} & = x_{FG}+ \gamma(1-x_{FG})
\end{aligned}
\label{eqfour}
\end{equation}
where $x_{FG}$ is the foreground region and $1-x_{FG}$ is the background region.
% which construct the foreground information and make the missing region and its inversion having a better fusion effect.
$\gamma$ is used to control the weight between foreground and background.
%At this time, the weight value of $W_{FB}$ is 1.0 in the foreground region and 0.5 in the background region.

The final loss for the Domain Embedded Generator is defined as follows:
\begin{equation}
\begin{aligned}
  \mathcal{L}_x^{rec} & =\mathop{\mathbb{E}}  [|| W_{FB}\otimes(x_{rec}  -x_{real}))||_{1}]
\end{aligned}
\label{eqfour}
\end{equation}
where $||\cdot||_1$ is the $L_1$ loss, $x_{rec}$ and $x_{real}$ represent the reconstructed image and the real image, respectively. And $\otimes$ is element-wise multiplying.
%$W_{FB}$ is a weight matrix about reinforcing the foreground region.

\begin{algorithm}[t]
   \caption{\emph{DEGNet} Algorithm. All experiments in this paper set $\alpha=0.0002$, $m=64$}
\label{alg:example}
\begin{algorithmic}[l]
\REQUIRE $\alpha$: learning rate, $m$: batch size, $W_{vae}$: Vae's parameters, $W_{u}$: FCN's parameters, $W_{g}$: global discriminator parameters, $W_p$: patch discriminator parameters, $V_{latent}$: sampled latent vector from Vae's encoder, $X_{prior}$: domain infromation about face, $D_p$: patch discriminator, $D_g$: global discriminator

\WHILE{$W_{vae}$ and $W_u$ have not converged}
\STATE $V_{latent}^{(i)}\sim sample\{x_{crop}^{(i)}\}_{i=1}^m$: sample a batch from the cropped data
\STATE $x_{m},x_{f},x_{l} \sim G\{V_{latent}^{(i)},V_{l}^{(i)}\}_{i=1}^m$: generate facial information from the Vae's decoder
\STATE $x_{rec}^{(i)} \sim G\{x_{crop}^{(i)},V_{latent}^{(i)}\}_{i=1}^m$: reconstructed images by FCN
\STATE $score_1 \sim D_p\{x_{rec}^{(i)}, x_{real}^{(i)}\}_{i=1}^m$: discriminate the distribution of $x^{(i)}$ real or fake
\STATE $score_2 \sim D_g\{x_{rec}^{(i)}, x_{real}^{(i)}\}_{i=1}^m$: discriminate the distribution of $x^{(i)}$ real or fake
\STATE Update $W_{vae}$ by $\mathcal{L}_{m}^{rec}$, $\mathcal{L}_{f}^{rec}$, $\mathcal{L}_{l}^{rec}$,
$\mathcal{L}_{f}^{lat}$,
$\mathcal{L}_{l}^{lat}$ and $\mathcal{L}_{m}^{lat}$
\STATE Update $W_u$ by $\mathcal{L}_{x}^{rec}$, $\mathcal{L}_{x}^{adv}$, $\mathcal{L}_{p}^{adv}$
\STATE Update $W_g$ and $W_p$ $\sim$ $\{score_1, score_2\}$
\ENDWHILE
\end{algorithmic}
\end{algorithm}

\textbf{Multi-model Discriminator Loss.}
The global adversarial loss and patch adversarial loss are described as follows:

\begin{equation}
\begin{aligned}
    \mathcal{L}^{adv-g}  =-\mathop{\mathbb{E}}[\log(1- D_{g}(x_{rec}))]-{\mathbb{E}}[log(D_{g}(x_{real}))]\\
   \mathcal{L}^{adv-p}  =-\mathop{\mathbb{E}}[\log(1- D_{p}(x_{rec}))]-{\mathbb{E}}[log(D_{P}(x_{real}))]
\end{aligned}
\label{eqadversial}
\end{equation}

where $D_{p}$ is the patch discriminator and $D_{g}$ is the global discriminator.

% The former distinguishes whether a patch of the image is from a real sample or a synthesized image. It can capture the local statistics and drive the generator to generate locally coherent face images. The g distinguishes whether an image is a real sample or a synthesized image. This can significantly improve adversarial training robustness and alleviate the transfer ability among the members of the ensembles in both untargeted and targeted modes.

\noindent \textbf{Total Loss.} The overall loss function of our model is defined by a weighted sum of the above loss functions:
\begin{equation}
\label{eqTotal}
\begin{aligned}
    \mathcal{L} & =\lambda_{f}^{rec}\mathcal{L}^{rec}_{f}+\lambda_{m}^{rec}\mathcal{L}^{rec}_{m}
    +\lambda_{l}^{rec}\mathcal{L}^{rec}_{l} \\
    & +\lambda^{lat}(\mathcal{L}^{lat}_{f}+\mathcal{L}^{lat}_{m}+\mathcal{L}^{lat}_{l})\\
    & +\lambda^{adv-g}\mathcal{L}^{adv-g}+\lambda^{adv-p}\mathcal{L}^{adv-p}
\end{aligned}
\end{equation}

The whole algorithm is  summarized in Algorithm~\ref{alg:example}.

\section{Experiments}

\subsection{Experiments Settings.}
\noindent \textbf{Datasets.} Our experiments are conducted on two human face datasets. 1) CelebA~\cite{liu2015faceattributes}, a Large-scale CelebFaces Attributes Dataset. 2) CelebA-HQ~\cite{karras2018progressive}, a high-quality  version of CelebA datset. We follow the official split for training, validating and testing (details in Table ~\ref{tb_dataset}).

\begin{table}[h]
\caption{Training and test splits of two datasets.}
 \label{tb_dataset}
%\vspace{-0.4cm}
% \setlength{\abovecaptionskip}{-0.01cm}
%\setlength{\belowcaptionskip}{-80pt}
    \centering
    \begin{tabular}{c|r|r|r}
        \hline
DataSet & $\#$Train & $\#$Val & $\#$Test\\
\hline
CelebA~\cite{liu2015faceattributes} & 162770  &19867  & 19962\\
CelebA-HQ~\cite{karras2018progressive} & 28000 & 1000 & 1000\\
\hline
\end{tabular}
%    \vspace{-0.6cm}
\end{table}

\noindent \textbf{Evaluation Metrics.} Four types of criteria are used to measure the performance of different methods: 1) Peak Signal to Noise Ratio (PSNR), which directly measures visibility of errors and gives you an average value; 2) Structural SIMilarity (SSIM), which measures the structural similarity of an image against a reference image; 3) Normalization Cross-Correlation (NCC), which has been commonly used to evaluate the degree of similarity between two compared images;  4) L1 loss, which can reﬂect the ability of models to reconstruct the original image.

\noindent \textbf{Pre-processing.} Our training dataset includes six types of images (see Fig.~\ref{fig_dataset}): 1) the original full-face image with $178\times218$ (Fig. ~\ref{fig_dataset}(a)), 2) cropped face image or landmarks (Fig.~\ref{fig_dataset}(b)), 3) face mask $x_{m}$ (Fig.~\ref{fig_dataset}(c)), 4) face part $x_{f}$ (Fig.~\ref{fig_dataset}(d)), 5) landmark image $x_{l}$ (Fig.~\ref{fig_dataset}(e)), we use face alignment detection interface~\cite{wang2018facial} to extract 68 facial landmarks properly from an original full-face image, and 6) foreground mask $x_{FG}$ (Fig.~\ref{fig_dataset}(f)).

The cropped face image (Fig.~\ref{fig_dataset}(b)) and face mask (Fig.~\ref{fig_dataset}(c)) are obtained by stretching the convex hull computed from the 41 landmarks in eyes nose and mouth.
To obtain the face part, we dilate the face mask by 3\% of the image width to ensure
that the mask borders are slightly inside the face contours and include the eyebrows inside the mask. Then the only face part image (Fig.~\ref{fig_dataset}(d)) is obtained by applying the face mask to the input image.
%Like FSNet \cite{Natsume2018FSNet}, we generated the landmark images (Figure \ref{fig_dataset}(e)).
Finally, the foreground mask (Fig.~\ref{fig_dataset}(f)) is detected using Baidu segementation API~\cite{baidu2019}.

For the Celeba dataset, all the face images are coarsely cropped from $178\times218$ to $178\times178$. Finally, the cropped images are resized to $128\times128$ in our experiment. For the Celeba-HQ dataset, the face image size is $256\times256$.

\noindent \textbf{Implementation Details.} All experiments are implemented using Tensorflow and Keras framework.

We use Adam optimizer~\cite{kingma2015adam} with an initial learning rate of 0.0002, $\beta_1=0.5$, $\beta_2=0.999$ and leave other parameters as Keras default. Our model uses batch size of 60 training with 80 epochs and sets $\lambda^{rec}_m=\lambda^{rec}_f =4,000$, $\lambda^{rec}_l=2,000$, $\lambda^{lat}=30$ and $\lambda^{dav-p}=30$. The $\gamma$ in the foreground and background weigh $W_{FB}$ (Eq. (\ref{eqfour})) is set to 0.5.

\begin{table}[t]
%\vspace{-0.4cm}
\setlength{\abovecaptionskip}{-0.01cm}
\caption{Comparison of the proposed DEGNet with the other advanced methods on CelebA. }
\label{tb_CelebA_qualitative}
    \centering
    \begin{tabular}{c|r|r|r}
        \hline
Method &  PSNR & SSIM & NCC\\
\hline
PM~\cite{barnes2009patchmatch} &22.0  &0.861  &0.931 \\
CE~\cite{pathak2016context} & 24.950 & 0.870 & 0.969\\
GLCIC~\cite{iizuka2017globally} &25.081  &0.875  &0.968\\
CA~\cite{yu2018generative} &23.897& 0.858& 0.962 \\
PICNet~\cite{zheng2019pluralistic} & 25.135 & 0.876 & 0.970 \\
PEN~\cite{yan2019PENnet} & 24.712 & 0.872 & 0.965 \\
DEGNet (Our)& \textbf{25.274} & \textbf{0.882} & \textbf{0.972}\\
\hline
    \end{tabular}
%    \vspace{-0.6cm}
\end{table}

\begin{table}[t]
%\vspace{-0.4cm}
\setlength{\abovecaptionskip}{-0.01cm}
\caption{Comparison of the proposed DEGNet with the other advanced methods on CelebA-HQ.}
\label{tb_CelebAHQ_qualitative}
    \centering
    \begin{tabular}{c|r|r|r|r}
        \hline
Method &  PSNR & SSIM & NCC & L1-loss\\
\hline
PM~\cite{barnes2009patchmatch} &23.190 &0.887 &0.952 & 52.02\\
CE~\cite{pathak2016context} & 26.076 & 0.885 & 0.978 & 44.13\\
GLCIC~\cite{iizuka2017globally} & 25.802 &0.886 &0.974  &44.35 \\
CA~\cite{yu2018generative} & 22.862 & 0.844 & 0.954 & 47.21\\
PICNet~\cite{zheng2019pluralistic} & 25.091 & 0.869 & 0.972 & 44.74 \\
PEN~\cite{yan2019PENnet} & 25.717 & 0.890 & 0.975 & 9.94 \\
DEGNet (Our)& \textbf{26.208} & \textbf{0.895} & \textbf{0.978}& \textbf{4.65}\\
\hline
    \end{tabular}
%    \vspace{-0.6cm}
\end{table}

\subsection{Comparison with Existing Work}
We compare our algorithm against existing works in two groups: general image inpainting methods and face inpainting methods separately:

\textbf{(1) General image inpainting methods:} The texture refinement methods PM~\cite{barnes2009patchmatch}, Context\_encoder (CE)~\cite{pathak2016context}, GLCIC~\cite{iizuka2017globally}, PEN~\cite{yan2019PENnet} and Generative\_inpainting (CA)~\cite{yu2018generative} that replacing the initially completed images and traditional methods which only using low-level features to complete image inpainting and  Pluralistic Network (PICNet)~\cite{zheng2019pluralistic}, noted that since PICNet can generate multiple outputs, we chose the top one best results based on its discriminator scores to compare. Both PICNet and CA methods require high-resolution training images in original papers, we report their results on the high-resolution CelebA-HQ dataset. For the PEN, we used their released model for testing\footnote{https://github.com/researchmm/PEN-Net-for-Inpainting}.

%And there is no public code available for the PENNet~\cite{yan2019PENnet}, we compare the best $L_1$ loss performance reported in their paper.

As presented in Table~\ref{tb_CelebA_qualitative} and Table~\ref{tb_CelebAHQ_qualitative},
our method, in general, achieves better performance than all the other methods, in terms of SSIM, PSNR, NCC, and $L_1$ loss. It is easy to see that our method outperforms the state-of-the-art in both PSNR and SSIM and
achieves the best generalization performance for large and different crops.
CA only achieve better results for high resolution training dataset, but get poor result for low resolution training dataset (see Fig.~\ref{intro_qualitative}(d)).

As presented in Fig.~\ref{fig_qualitative}, the first six lines of inpainting results are from the Celeba, and others are from the Celeba-HQ. When the missing area is quite different from the surrounding environment, we find that PM method does not paint the whole face, and the CE method has better performance in some frontal faces but when the missing area contains some background information besides the face, it is a high possibility for CE to produce blurry even distorted face. And the faces generated by GLCIC are also blurry. The PIC can complete clear faces but the faces are not harmonious. this is because PIC is to produce a clearer image by enhancing the constraint capability of the discriminator, which destroy the structural consistency of the image and result in distortion of the image. { The PEN has better performance in the Celeba-HQ dataset, but it performs poorly in the Celeba dataset.} In DEGNet, we produce clear and harmonious face inpainting by keeping a balance between rec loss and adv loss.

\afterpage{%

\begin{figure*}[t]
    %\vskip 0.2in
    \begin{center}
        \begin{minipage}[t]{0.10\textwidth}
         \includegraphics[width=\linewidth]{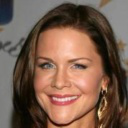}
        \end{minipage}
        \begin{minipage}[t]{0.10\textwidth}
            \includegraphics[width=\linewidth]{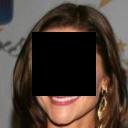}
        \end{minipage}
        \begin{minipage}[t]{0.10\textwidth}
            \includegraphics[width=\linewidth]{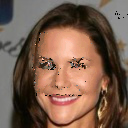}
        \end{minipage}
        \begin{minipage}[t]{0.10\textwidth}
            \includegraphics[width=\linewidth]{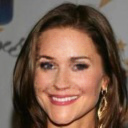}
        \end{minipage}
        \begin{minipage}[t]{0.10\textwidth}
            \includegraphics[width=\linewidth]{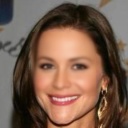}
        \end{minipage}
        \begin{minipage}[t]{0.10\textwidth}
            \includegraphics[width=\linewidth]{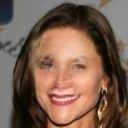}
        \end{minipage}
        \begin{minipage}[t]{0.10\textwidth}
            \includegraphics[width=\linewidth]{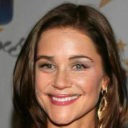}
        \end{minipage}
        \begin{minipage}[t]{0.10\textwidth}
            \includegraphics[width=\linewidth]{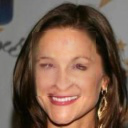}
       \end{minipage}
        \begin{minipage}[t]{0.10\textwidth}
            \includegraphics[width=\linewidth]{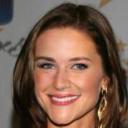}

       \end{minipage}

       \begin{minipage}[t]{0.10\textwidth}
            \includegraphics[width=\linewidth]{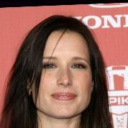}
        \end{minipage}
        \begin{minipage}[t]{0.10\textwidth}
            \includegraphics[width=\linewidth]{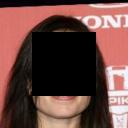}
        \end{minipage}
        \begin{minipage}[t]{0.10\textwidth}
            \includegraphics[width=\linewidth]{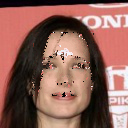}
        \end{minipage}
        \begin{minipage}[t]{0.10\textwidth}
            \includegraphics[width=\linewidth]{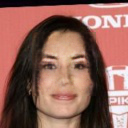}
        \end{minipage}
        \begin{minipage}[t]{0.10\textwidth}
            \includegraphics[width=\linewidth]{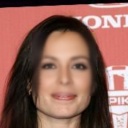}
        \end{minipage}
        \begin{minipage}[t]{0.10\textwidth}
            \includegraphics[width=\linewidth]{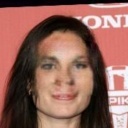}
        \end{minipage}
        \begin{minipage}[t]{0.10\textwidth}
            \includegraphics[width=\linewidth]{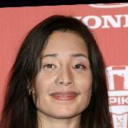}
        \end{minipage}
        \begin{minipage}[t]{0.10\textwidth}
            \includegraphics[width=\linewidth]{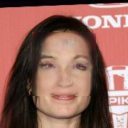}
        \end{minipage}
        \begin{minipage}[t]{0.10\textwidth}
            \includegraphics[width=\linewidth]{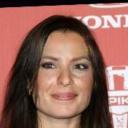}
        \end{minipage}

        \begin{minipage}[t]{0.10\textwidth}
            \includegraphics[width=\linewidth]{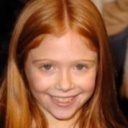}
        \end{minipage}
        \begin{minipage}[t]{0.10\textwidth}
            \includegraphics[width=\linewidth]{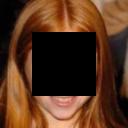}
        \end{minipage}
        \begin{minipage}[t]{0.10\textwidth}
            \includegraphics[width=\linewidth]{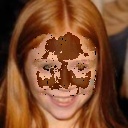}
        \end{minipage}
        \begin{minipage}[t]{0.10\textwidth}
            \includegraphics[width=\linewidth]{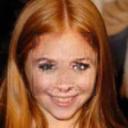}
        \end{minipage}
        \begin{minipage}[t]{0.10\textwidth}
            \includegraphics[width=\linewidth]{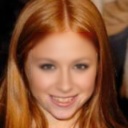}
        \end{minipage}
        \begin{minipage}[t]{0.10\textwidth}
            \includegraphics[width=\linewidth]{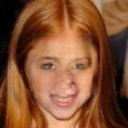}
        \end{minipage}
        \begin{minipage}[t]{0.10\textwidth}
            \includegraphics[width=\linewidth]{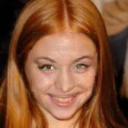}
        \end{minipage}
        \begin{minipage}[t]{0.10\textwidth}
            \includegraphics[width=\linewidth]{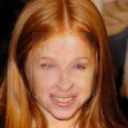}
        \end{minipage}
        \begin{minipage}[t]{0.10\textwidth}
            \includegraphics[width=\linewidth]{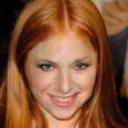}
        \end{minipage}

        \begin{minipage}[t]{0.10\textwidth}
            \includegraphics[width=\linewidth]{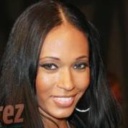}
        \end{minipage}
        \begin{minipage}[t]{0.10\textwidth}
            \includegraphics[width=\linewidth]{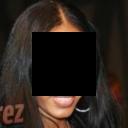}
        \end{minipage}
        \begin{minipage}[t]{0.10\textwidth}
            \includegraphics[width=\linewidth]{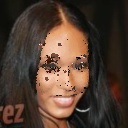}
        \end{minipage}
        \begin{minipage}[t]{0.10\textwidth}
            \includegraphics[width=\linewidth]{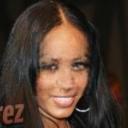}
        \end{minipage}
        \begin{minipage}[t]{0.10\textwidth}
            \includegraphics[width=\linewidth]{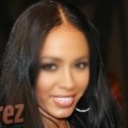}
        \end{minipage}
        \begin{minipage}[t]{0.10\textwidth}
            \includegraphics[width=\linewidth]{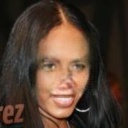}
        \end{minipage}
        \begin{minipage}[t]{0.10\textwidth}
            \includegraphics[width=\linewidth]{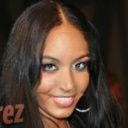}
        \end{minipage}
        \begin{minipage}[t]{0.10\textwidth}
            \includegraphics[width=\linewidth]{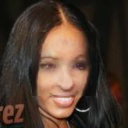}
        \end{minipage}
        \begin{minipage}[t]{0.10\textwidth}
            \includegraphics[width=\linewidth]{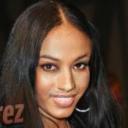}
        \end{minipage}

        \begin{minipage}[t]{0.10\textwidth}
            \includegraphics[width=\linewidth]{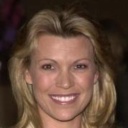}
        \end{minipage}
        \begin{minipage}[t]{0.10\textwidth}
            \includegraphics[width=\linewidth]{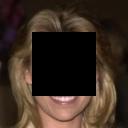}
        \end{minipage}
        \begin{minipage}[t]{0.10\textwidth}
            \includegraphics[width=\linewidth]{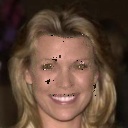}
        \end{minipage}
        \begin{minipage}[t]{0.10\textwidth}
            \includegraphics[width=\linewidth]{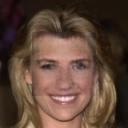}
        \end{minipage}
        \begin{minipage}[t]{0.10\textwidth}
            \includegraphics[width=\linewidth]{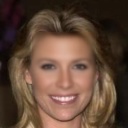}
        \end{minipage}
        \begin{minipage}[t]{0.10\textwidth}
            \includegraphics[width=\linewidth]{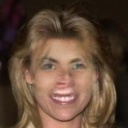}
        \end{minipage}
        \begin{minipage}[t]{0.10\textwidth}
            \includegraphics[width=\linewidth]{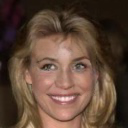}
        \end{minipage}
        \begin{minipage}[t]{0.10\textwidth}
            \includegraphics[width=\linewidth]{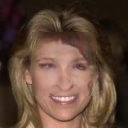}
        \end{minipage}
        \begin{minipage}[t]{0.10\textwidth}
            \includegraphics[width=\linewidth]{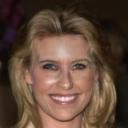}
        \end{minipage}

        \begin{minipage}[t]{0.10\textwidth}
            \includegraphics[width=\linewidth]{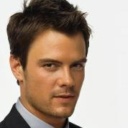}
        \end{minipage}
        \begin{minipage}[t]{0.10\textwidth}
            \includegraphics[width=\linewidth]{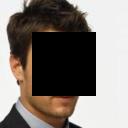}
        \end{minipage}
        \begin{minipage}[t]{0.10\textwidth}
            \includegraphics[width=\linewidth]{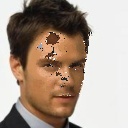}
        \end{minipage}
        \begin{minipage}[t]{0.10\textwidth}
            \includegraphics[width=\linewidth]{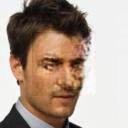}
        \end{minipage}
        \begin{minipage}[t]{0.10\textwidth}
            \includegraphics[width=\linewidth]{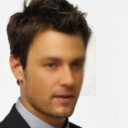}
        \end{minipage}
        \begin{minipage}[t]{0.10\textwidth}
            \includegraphics[width=\linewidth]{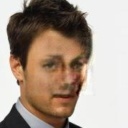}
        \end{minipage}
        \begin{minipage}[t]{0.10\textwidth}
            \includegraphics[width=\linewidth]{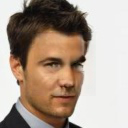}
        \end{minipage}
        \begin{minipage}[t]{0.10\textwidth}
            \includegraphics[width=\linewidth]{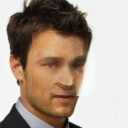}
        \end{minipage}
        \begin{minipage}[t]{0.10\textwidth}
            \includegraphics[width=\linewidth]{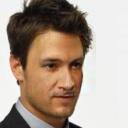}
        \end{minipage}

        \begin{minipage}[t]{0.10\textwidth}
            \includegraphics[width=\linewidth]{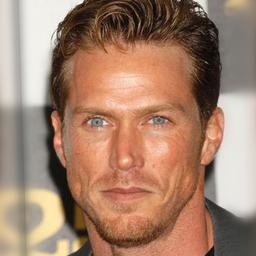}
        \end{minipage}
        \begin{minipage}[t]{0.10\textwidth}
            \includegraphics[width=\linewidth]{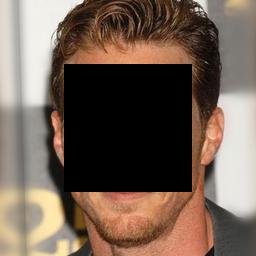}
        \end{minipage}
        \begin{minipage}[t]{0.10\textwidth}
            \includegraphics[width=\linewidth]{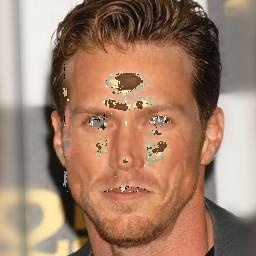}
        \end{minipage}
        \begin{minipage}[t]{0.10\textwidth}
            \includegraphics[width=\linewidth]{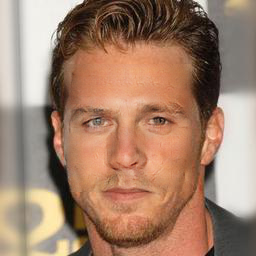}
        \end{minipage}
        \begin{minipage}[t]{0.10\textwidth}
            \includegraphics[width=\linewidth]{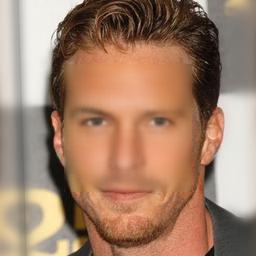}
        \end{minipage}
        \begin{minipage}[t]{0.10\textwidth}
            \includegraphics[width=\linewidth]{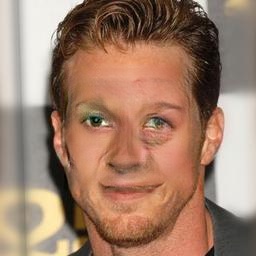}
        \end{minipage}
        \begin{minipage}[t]{0.10\textwidth}
            \includegraphics[width=\linewidth]{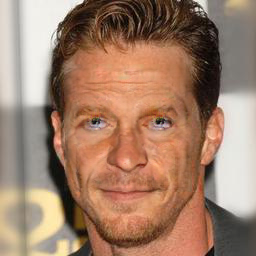}
        \end{minipage}
        \begin{minipage}[t]{0.10\textwidth}
            \includegraphics[width=\linewidth]{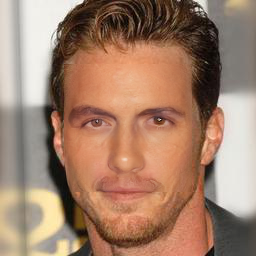}
        \end{minipage}
        \begin{minipage}[t]{0.10\textwidth}
            \includegraphics[width=\linewidth]{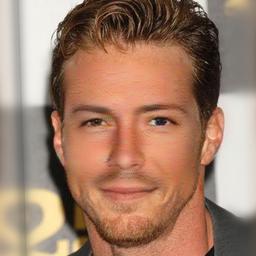}
        \end{minipage}

        \begin{minipage}[t]{0.10\textwidth}
            \includegraphics[width=\linewidth]{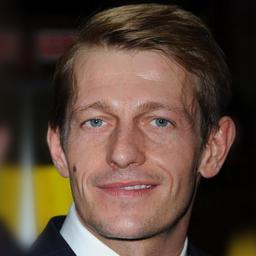}
        \end{minipage}
        \begin{minipage}[t]{0.10\textwidth}
            \includegraphics[width=\linewidth]{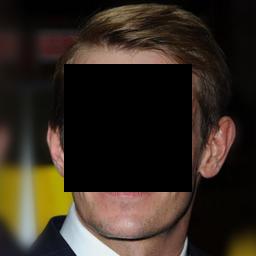}
        \end{minipage}
        \begin{minipage}[t]{0.10\textwidth}
            \includegraphics[width=\linewidth]{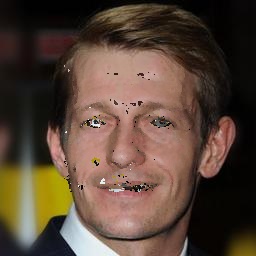}
        \end{minipage}
        \begin{minipage}[t]{0.10\textwidth}
            \includegraphics[width=\linewidth]{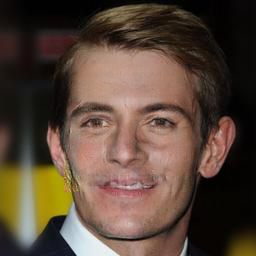}
        \end{minipage}
        \begin{minipage}[t]{0.10\textwidth}
            \includegraphics[width=\linewidth]{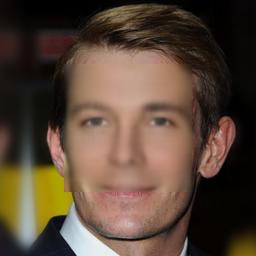}
        \end{minipage}
        \begin{minipage}[t]{0.10\textwidth}
            \includegraphics[width=\linewidth]{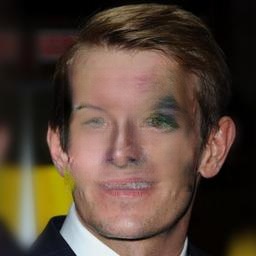}
        \end{minipage}
        \begin{minipage}[t]{0.10\textwidth}
            \includegraphics[width=\linewidth]{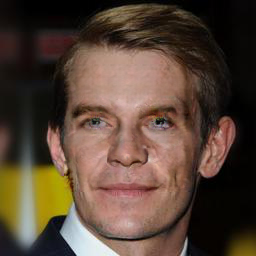}
        \end{minipage}
        \begin{minipage}[t]{0.10\textwidth}
            \includegraphics[width=\linewidth]{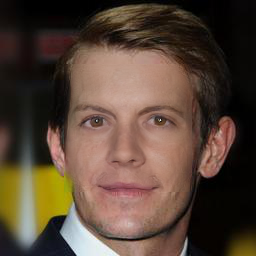}
        \end{minipage}
        \begin{minipage}[t]{0.10\textwidth}
            \includegraphics[width=\linewidth]{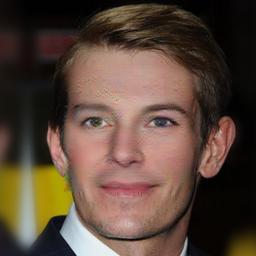}
        \end{minipage}

        \begin{minipage}[t]{0.10\textwidth}
            \includegraphics[width=\linewidth]{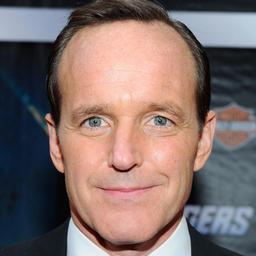}
        \end{minipage}
        \begin{minipage}[t]{0.10\textwidth}
            \includegraphics[width=\linewidth]{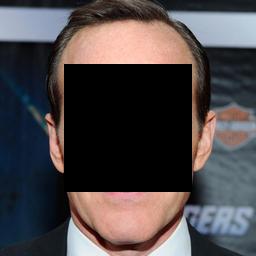}
        \end{minipage}
        \begin{minipage}[t]{0.10\textwidth}
            \includegraphics[width=\linewidth]{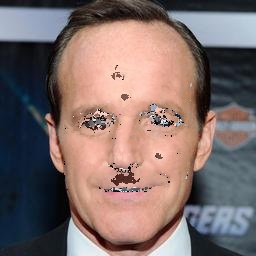}
        \end{minipage}
        \begin{minipage}[t]{0.10\textwidth}
            \includegraphics[width=\linewidth]{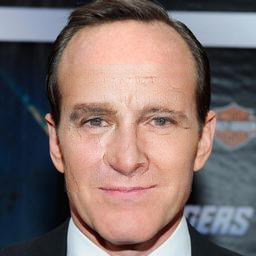}
        \end{minipage}
        \begin{minipage}[t]{0.10\textwidth}
            \includegraphics[width=\linewidth]{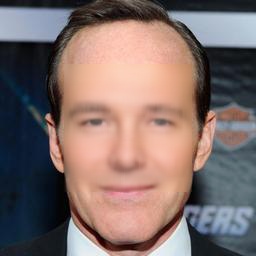}
        \end{minipage}
        \begin{minipage}[t]{0.10\textwidth}
            \includegraphics[width=\linewidth]{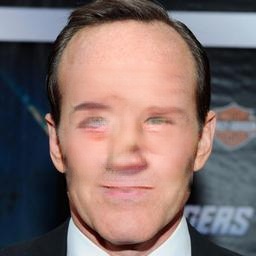}
        \end{minipage}
        \begin{minipage}[t]{0.10\textwidth}
            \includegraphics[width=\linewidth]{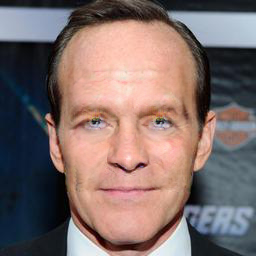}
        \end{minipage}
        \begin{minipage}[t]{0.10\textwidth}
            \includegraphics[width=\linewidth]{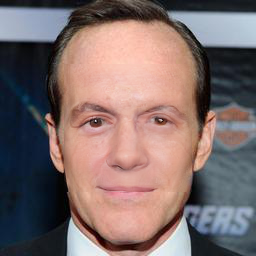}
        \end{minipage}
        \begin{minipage}[t]{0.10\textwidth}
            \includegraphics[width=\linewidth]{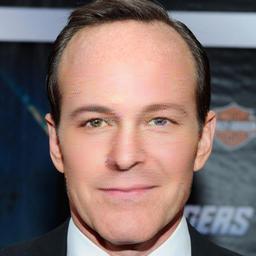}
        \end{minipage}

        \begin{minipage}[t]{0.10\textwidth}
            \centerline{\includegraphics[width=\linewidth]{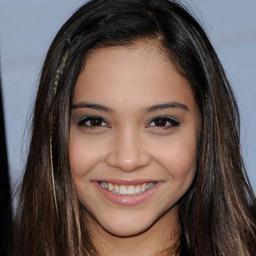}}
            \centerline{(a) Original}
        \end{minipage}
        \begin{minipage}[t]{0.10\textwidth}
            \centerline{\includegraphics[width=\linewidth]{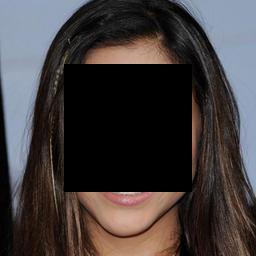}}
            \centerline{(b) Cropped}
        \end{minipage}
        \begin{minipage}[t]{0.10\textwidth}
            \centerline{\includegraphics[width=\linewidth]{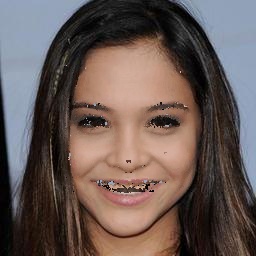}}
            \centerline{(c) PM}
        \end{minipage}
        \begin{minipage}[t]{0.10\textwidth}
            \centerline{\includegraphics[width=\linewidth]{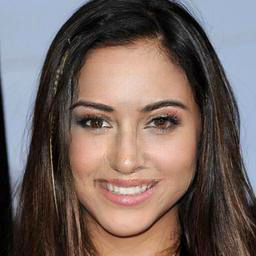}}
            \centerline{(d) CE}
        \end{minipage}
        \begin{minipage}[t]{0.10\textwidth}
            \centerline{\includegraphics[width=\linewidth]{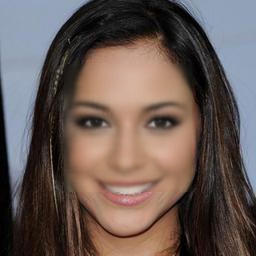}}
            \centerline{(e) GLCIC}
        \end{minipage}
        \begin{minipage}[t]{0.10\textwidth}
            \centerline{\includegraphics[width=\linewidth]{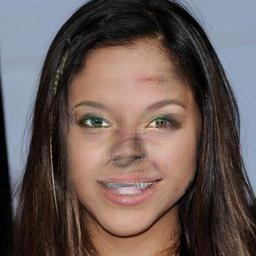}}
            \centerline{(f) CA}
        \end{minipage}
        \begin{minipage}[t]{0.10\textwidth}
            \centerline{\includegraphics[width=\linewidth]{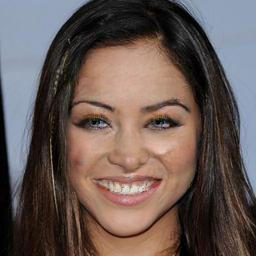}}
            \centerline{(g) PICNet}
        \end{minipage}
        \begin{minipage}[t]{0.10\textwidth}
            \centerline{\includegraphics[width=\linewidth]{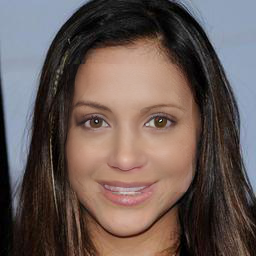}}
            \centerline{(h) PEN}
        \end{minipage}
        \begin{minipage}[t]{0.10\textwidth}
            \centerline{\includegraphics[width=\linewidth]{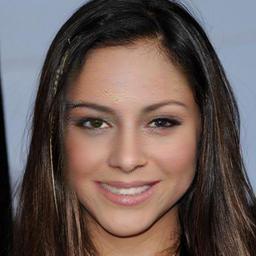}}
            \centerline{(I) DEGNet}
        \end{minipage}

    \end{center}
    \caption{Qualitative results of different inpainting methods on the CelebA dataset (the first six rows) and CelebA-HQ dataset (the rest rows).}
    \label{fig_qualitative}
    %\vskip -0.2in
\end{figure*}

\begin{figure*}[t]
    %\vskip 0.2in
    \begin{center}

        \begin{minipage}[t]{0.10\textwidth}
            \includegraphics[width=\linewidth]{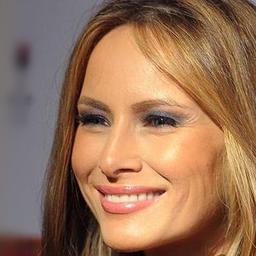}
        \end{minipage}
        \begin{minipage}[t]{0.10\textwidth}
            \includegraphics[width=\linewidth]{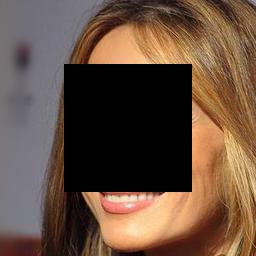}
        \end{minipage}
        \begin{minipage}[t]{0.10\textwidth}
            \includegraphics[width=\linewidth]{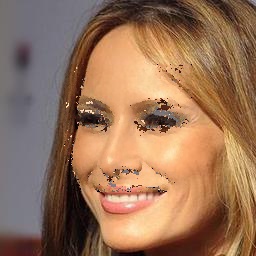}
        \end{minipage}
        \begin{minipage}[t]{0.10\textwidth}
            \includegraphics[width=\linewidth]{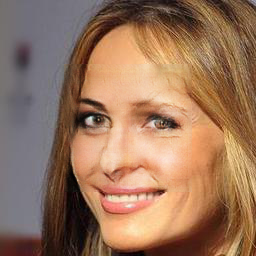}
        \end{minipage}
        \begin{minipage}[t]{0.10\textwidth}
            \includegraphics[width=\linewidth]{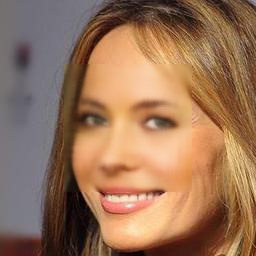}
        \end{minipage}
        \begin{minipage}[t]{0.10\textwidth}
            \includegraphics[width=\linewidth]{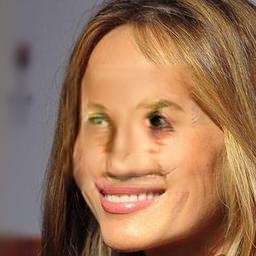}
        \end{minipage}
        \begin{minipage}[t]{0.10\textwidth}
            \includegraphics[width=\linewidth]{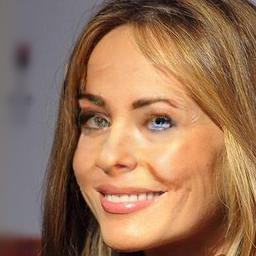}
        \end{minipage}
        \begin{minipage}[t]{0.10\textwidth}
            \includegraphics[width=\linewidth]{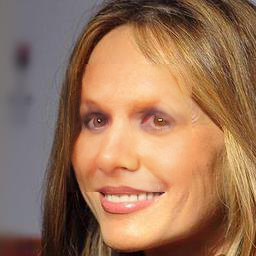}
        \end{minipage}
        \begin{minipage}[t]{0.10\textwidth}
            \includegraphics[width=\linewidth]{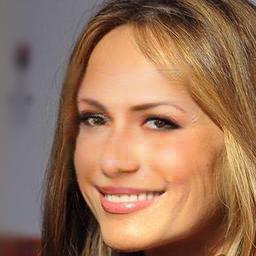}
        \end{minipage}

        \begin{minipage}[t]{0.10\textwidth}
            \includegraphics[width=\linewidth]{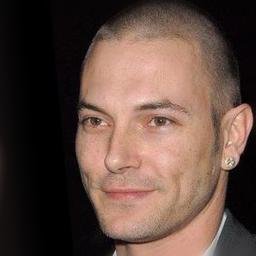}
        \end{minipage}
        \begin{minipage}[t]{0.10\textwidth}
            \includegraphics[width=\linewidth]{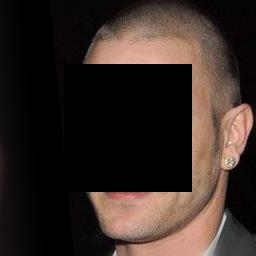}
        \end{minipage}
        \begin{minipage}[t]{0.10\textwidth}
            \includegraphics[width=\linewidth]{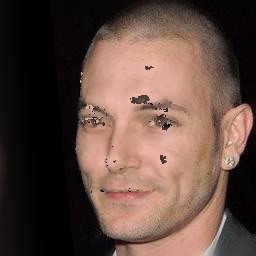}
        \end{minipage}
        \begin{minipage}[t]{0.10\textwidth}
            \includegraphics[width=\linewidth]{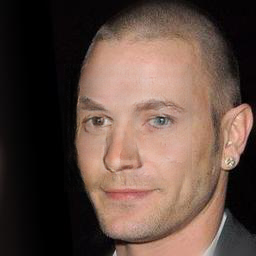}
        \end{minipage}
        \begin{minipage}[t]{0.10\textwidth}
            \includegraphics[width=\linewidth]{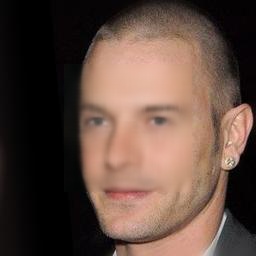}
        \end{minipage}
        \begin{minipage}[t]{0.10\textwidth}
            \includegraphics[width=\linewidth]{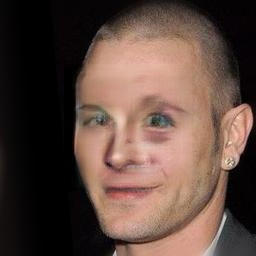}
        \end{minipage}
        \begin{minipage}[t]{0.10\textwidth}
            \includegraphics[width=\linewidth]{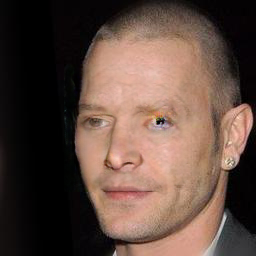}
        \end{minipage}
        \begin{minipage}[t]{0.10\textwidth}
            \includegraphics[width=\linewidth]{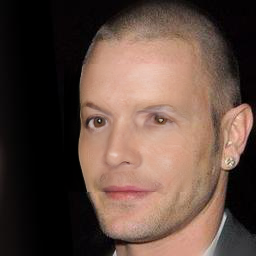}
        \end{minipage}
        \begin{minipage}[t]{0.10\textwidth}
            \includegraphics[width=\linewidth]{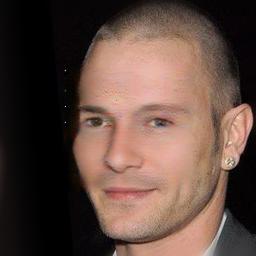}
        \end{minipage}

        \begin{minipage}[t]{0.10\textwidth}
            \includegraphics[width=\linewidth]{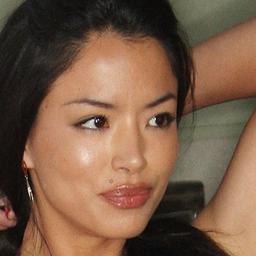}
        \end{minipage}
        \begin{minipage}[t]{0.10\textwidth}
            \includegraphics[width=\linewidth]{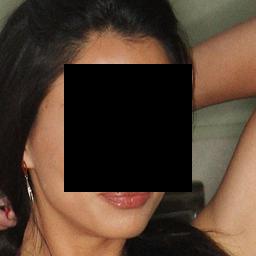}
        \end{minipage}
        \begin{minipage}[t]{0.10\textwidth}
            \includegraphics[width=\linewidth]{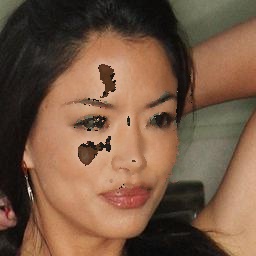}
        \end{minipage}
        \begin{minipage}[t]{0.10\textwidth}
            \includegraphics[width=\linewidth]{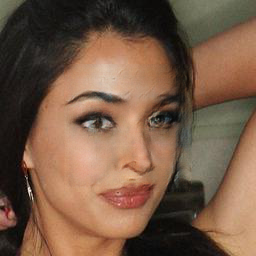}
        \end{minipage}
        \begin{minipage}[t]{0.10\textwidth}
            \includegraphics[width=\linewidth]{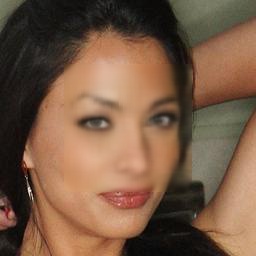}
        \end{minipage}
        \begin{minipage}[t]{0.10\textwidth}
            \includegraphics[width=\linewidth]{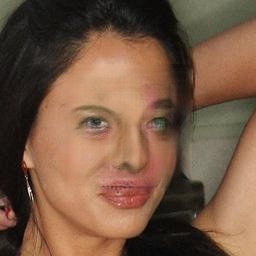}
        \end{minipage}
        \begin{minipage}[t]{0.10\textwidth}
            \includegraphics[width=\linewidth]{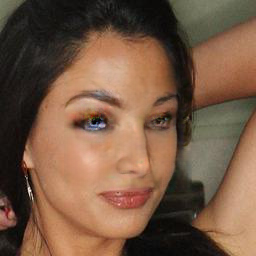}
        \end{minipage}
        \begin{minipage}[t]{0.10\textwidth}
            \includegraphics[width=\linewidth]{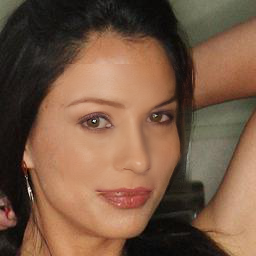}
        \end{minipage}
        \begin{minipage}[t]{0.10\textwidth}
            \includegraphics[width=\linewidth]{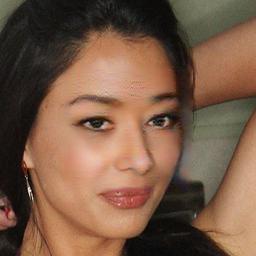}
        \end{minipage}

        \begin{minipage}[t]{0.10\textwidth}
            \includegraphics[width=\linewidth]{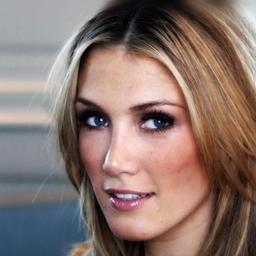}
        \end{minipage}
        \begin{minipage}[t]{0.10\textwidth}
            \includegraphics[width=\linewidth]{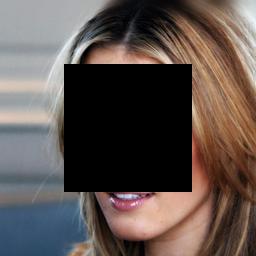}
        \end{minipage}
        \begin{minipage}[t]{0.10\textwidth}
            \includegraphics[width=\linewidth]{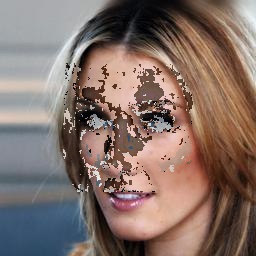}
        \end{minipage}
        \begin{minipage}[t]{0.10\textwidth}
            \includegraphics[width=\linewidth]{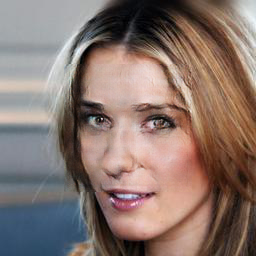}
        \end{minipage}
        \begin{minipage}[t]{0.10\textwidth}
            \includegraphics[width=\linewidth]{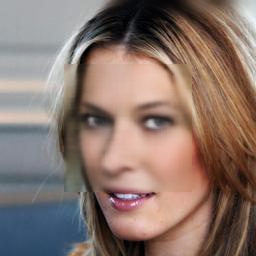}
        \end{minipage}
        \begin{minipage}[t]{0.10\textwidth}
            \includegraphics[width=\linewidth]{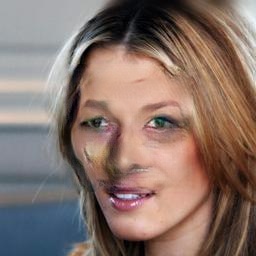}
        \end{minipage}
        \begin{minipage}[t]{0.10\textwidth}
            \includegraphics[width=\linewidth]{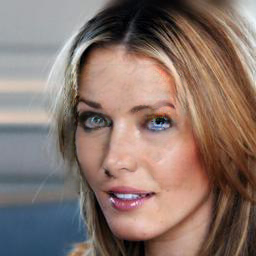}
        \end{minipage}
        \begin{minipage}[t]{0.10\textwidth}
            \includegraphics[width=\linewidth]{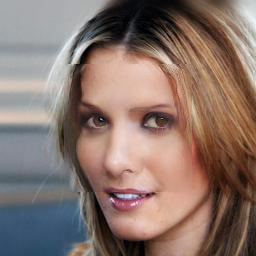}
        \end{minipage}
        \begin{minipage}[t]{0.10\textwidth}
            \includegraphics[width=\linewidth]{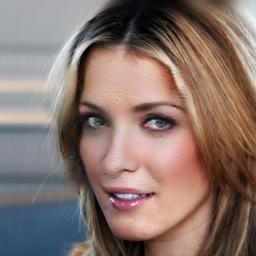}
        \end{minipage}

        \begin{minipage}[t]{0.10\textwidth}
            \includegraphics[width=\linewidth]{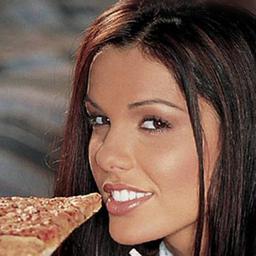}
        \end{minipage}
        \begin{minipage}[t]{0.10\textwidth}
            \includegraphics[width=\linewidth]{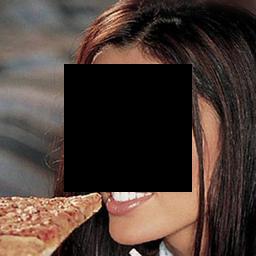}
        \end{minipage}
        \begin{minipage}[t]{0.10\textwidth}
            \includegraphics[width=\linewidth]{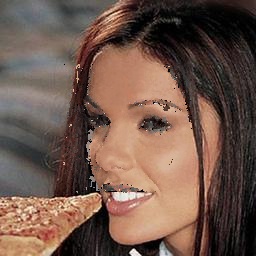}
        \end{minipage}
        \begin{minipage}[t]{0.10\textwidth}
            \includegraphics[width=\linewidth]{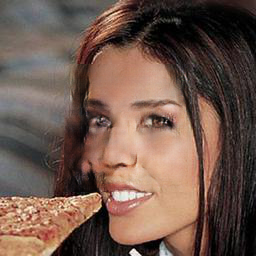}
        \end{minipage}
        \begin{minipage}[t]{0.10\textwidth}
            \includegraphics[width=\linewidth]{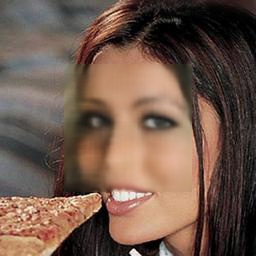}
        \end{minipage}
        \begin{minipage}[t]{0.10\textwidth}
            \includegraphics[width=\linewidth]{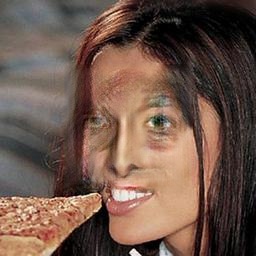}
        \end{minipage}
        \begin{minipage}[t]{0.10\textwidth}
            \includegraphics[width=\linewidth]{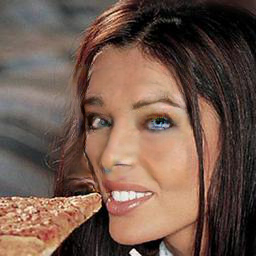}
        \end{minipage}
        \begin{minipage}[t]{0.10\textwidth}
            \includegraphics[width=\linewidth]{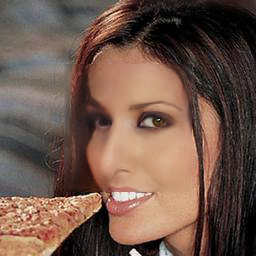}
        \end{minipage}
        \begin{minipage}[t]{0.10\textwidth}
            \includegraphics[width=\linewidth]{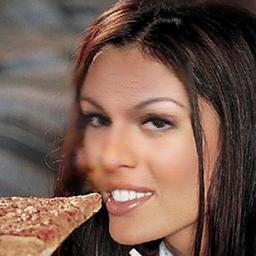}
        \end{minipage}

        \begin{minipage}[t]{0.10\textwidth}
            \includegraphics[width=\linewidth]{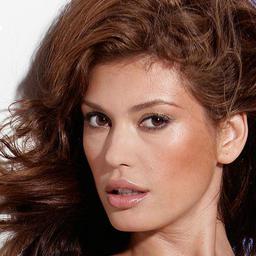}
        \end{minipage}
        \begin{minipage}[t]{0.10\textwidth}
            \includegraphics[width=\linewidth]{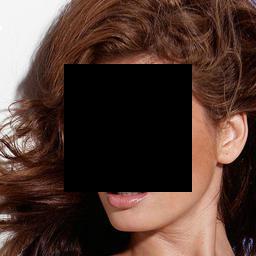}
        \end{minipage}
        \begin{minipage}[t]{0.10\textwidth}
            \includegraphics[width=\linewidth]{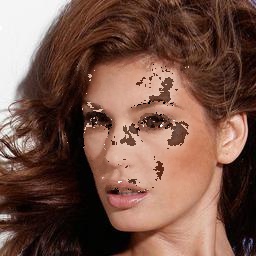}
        \end{minipage}
        \begin{minipage}[t]{0.10\textwidth}
            \includegraphics[width=\linewidth]{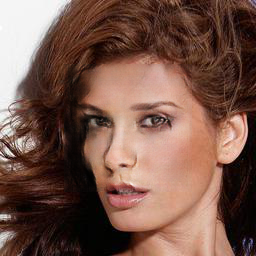}
        \end{minipage}
        \begin{minipage}[t]{0.10\textwidth}
            \includegraphics[width=\linewidth]{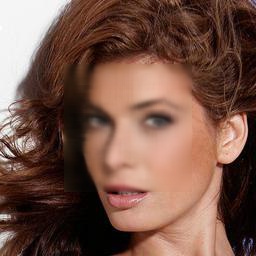}
        \end{minipage}
        \begin{minipage}[t]{0.10\textwidth}
            \includegraphics[width=\linewidth]{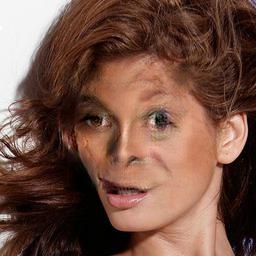}
        \end{minipage}
        \begin{minipage}[t]{0.10\textwidth}
            \includegraphics[width=\linewidth]{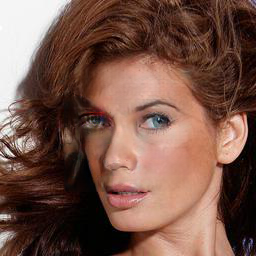}
        \end{minipage}
        \begin{minipage}[t]{0.10\textwidth}
            \includegraphics[width=\linewidth]{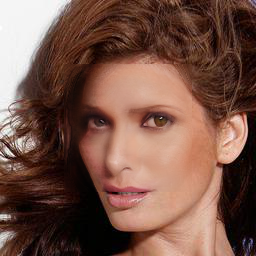}
        \end{minipage}
        \begin{minipage}[t]{0.10\textwidth}
            \includegraphics[width=\linewidth]{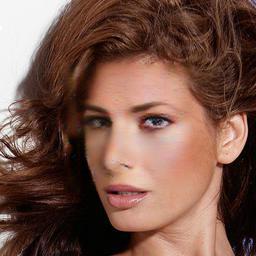}
        \end{minipage}

        \begin{minipage}[t]{0.10\textwidth}
            \includegraphics[width=\linewidth]{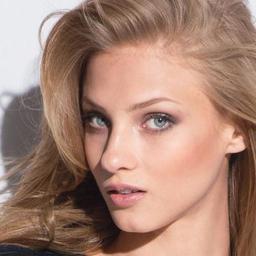}
        \end{minipage}
        \begin{minipage}[t]{0.10\textwidth}
            \includegraphics[width=\linewidth]{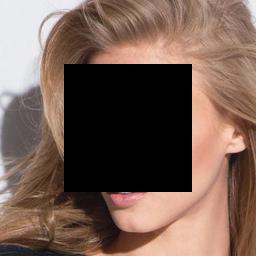}
        \end{minipage}
        \begin{minipage}[t]{0.10\textwidth}
            \includegraphics[width=\linewidth]{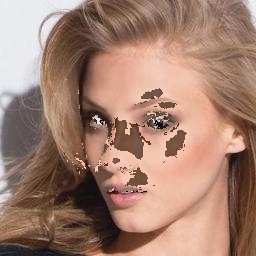}
        \end{minipage}
        \begin{minipage}[t]{0.10\textwidth}
            \includegraphics[width=\linewidth]{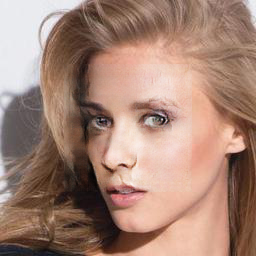}
        \end{minipage}
        \begin{minipage}[t]{0.10\textwidth}
            \includegraphics[width=\linewidth]{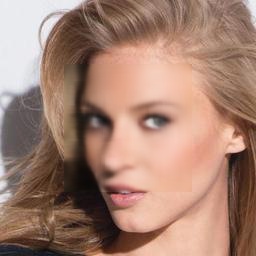}
        \end{minipage}
        \begin{minipage}[t]{0.10\textwidth}
            \includegraphics[width=\linewidth]{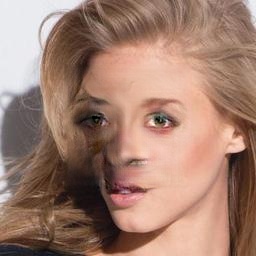}
        \end{minipage}
        \begin{minipage}[t]{0.10\textwidth}
            \includegraphics[width=\linewidth]{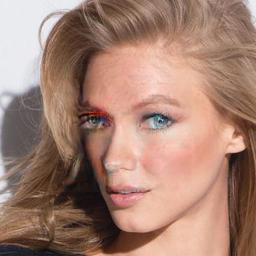}
        \end{minipage}
        \begin{minipage}[t]{0.10\textwidth}
            \includegraphics[width=\linewidth]{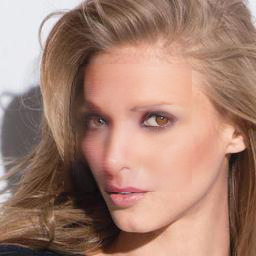}
        \end{minipage}
        \begin{minipage}[t]{0.10\textwidth}
            \includegraphics[width=\linewidth]{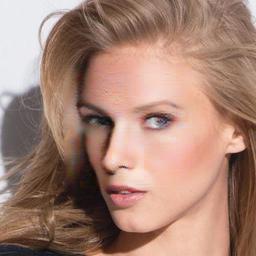}
        \end{minipage}

        \begin{minipage}[t]{0.10\textwidth}
            \includegraphics[width=\linewidth]{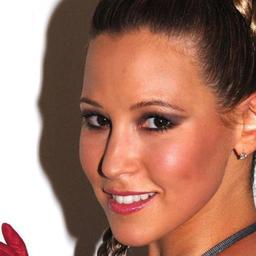}
        \end{minipage}
        \begin{minipage}[t]{0.10\textwidth}
            \includegraphics[width=\linewidth]{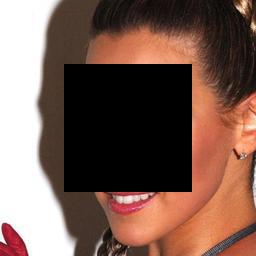}
        \end{minipage}
        \begin{minipage}[t]{0.10\textwidth}
            \includegraphics[width=\linewidth]{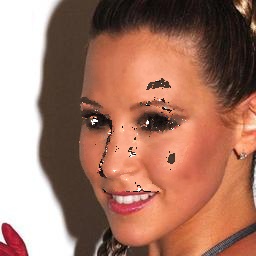}
        \end{minipage}
        \begin{minipage}[t]{0.10\textwidth}
            \includegraphics[width=\linewidth]{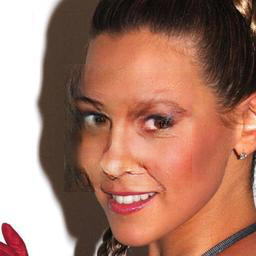}
        \end{minipage}
        \begin{minipage}[t]{0.10\textwidth}
            \includegraphics[width=\linewidth]{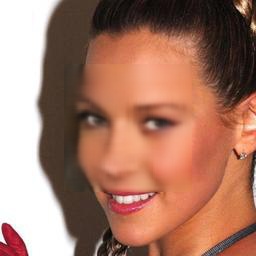}
        \end{minipage}
        \begin{minipage}[t]{0.10\textwidth}
            \includegraphics[width=\linewidth]{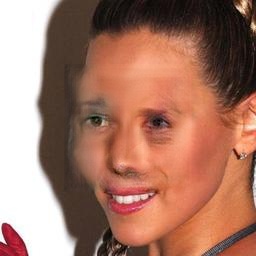}
        \end{minipage}
        \begin{minipage}[t]{0.10\textwidth}
            \includegraphics[width=\linewidth]{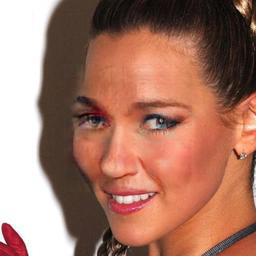}
        \end{minipage}
        \begin{minipage}[t]{0.10\textwidth}
            \includegraphics[width=\linewidth]{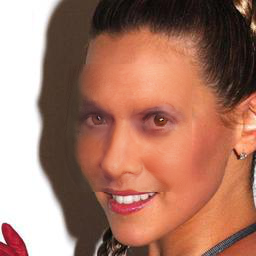}
        \end{minipage}
        \begin{minipage}[t]{0.10\textwidth}
            \includegraphics[width=\linewidth]{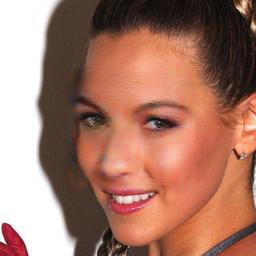}
        \end{minipage}

        \begin{minipage}[t]{0.10\textwidth}
            \includegraphics[width=\linewidth]{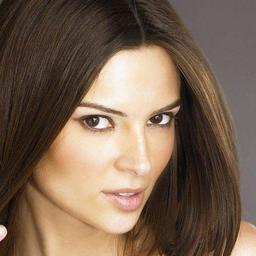}
        \end{minipage}
        \begin{minipage}[t]{0.10\textwidth}
            \includegraphics[width=\linewidth]{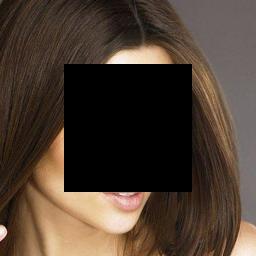}
        \end{minipage}
        \begin{minipage}[t]{0.10\textwidth}
            \includegraphics[width=\linewidth]{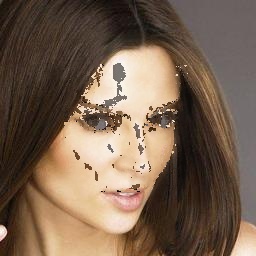}
        \end{minipage}
        \begin{minipage}[t]{0.10\textwidth}
            \includegraphics[width=\linewidth]{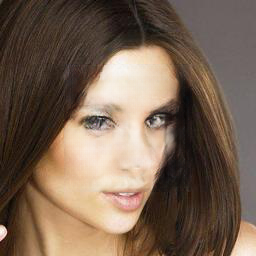}
        \end{minipage}
        \begin{minipage}[t]{0.10\textwidth}
            \includegraphics[width=\linewidth]{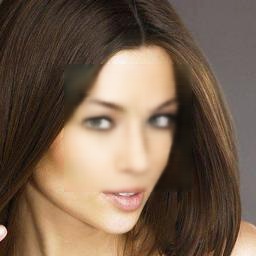}
        \end{minipage}
        \begin{minipage}[t]{0.10\textwidth}
            \includegraphics[width=\linewidth]{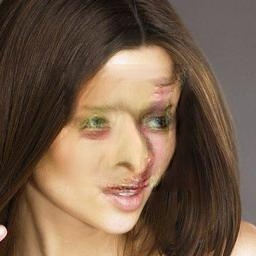}
        \end{minipage}
        \begin{minipage}[t]{0.10\textwidth}
            \includegraphics[width=\linewidth]{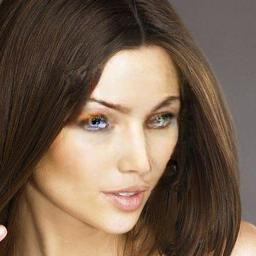}
        \end{minipage}
        \begin{minipage}[t]{0.10\textwidth}
            \includegraphics[width=\linewidth]{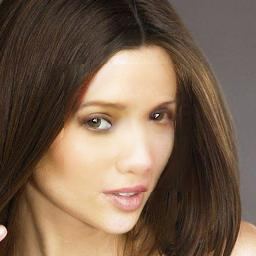}
        \end{minipage}
        \begin{minipage}[t]{0.10\textwidth}
            \includegraphics[width=\linewidth]{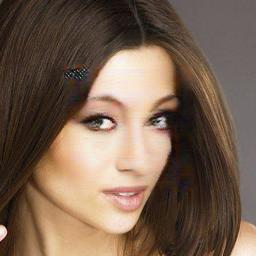}
        \end{minipage}

        \begin{minipage}[t]{0.10\textwidth}
            \centerline{\includegraphics[width=\linewidth]{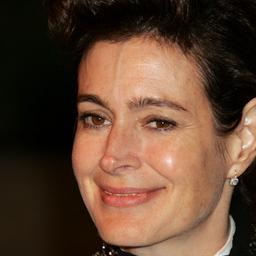}}
            \centerline{(a) Original}
        \end{minipage}
        \begin{minipage}[t]{0.10\textwidth}
            \centerline{\includegraphics[width=\linewidth]{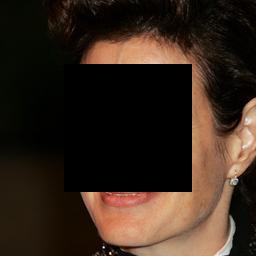}}
            \centerline{(b) Cropped}
        \end{minipage}
        \begin{minipage}[t]{0.10\textwidth}
            \centerline{\includegraphics[width=\linewidth]{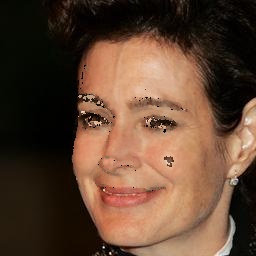}}
            \centerline{(c) PM}
        \end{minipage}
        \begin{minipage}[t]{0.10\textwidth}
            \centerline{\includegraphics[width=\linewidth]{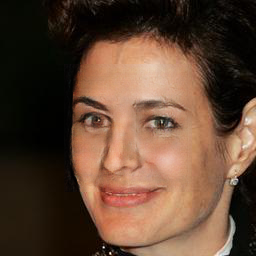}}
            \centerline{(d) CE}
        \end{minipage}
        \begin{minipage}[t]{0.10\textwidth}
            \centerline{\includegraphics[width=\linewidth]{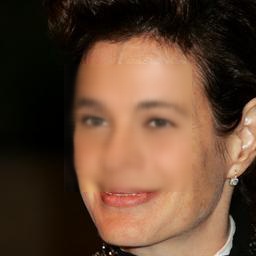}}
            \centerline{(e) GLCIC}
        \end{minipage}
        \begin{minipage}[t]{0.10\textwidth}
            \centerline{\includegraphics[width=\linewidth]{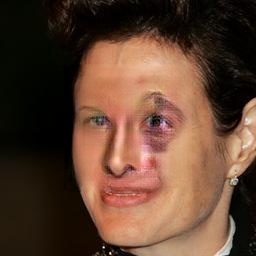}}
            \centerline{(f) CA}
        \end{minipage}
        \begin{minipage}[t]{0.10\textwidth}
            \centerline{\includegraphics[width=\linewidth]{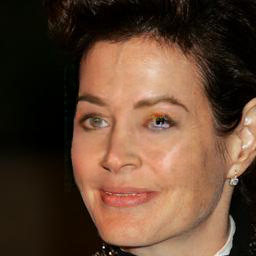}}
            \centerline{(g) PICNet}
        \end{minipage}
        \begin{minipage}[t]{0.10\textwidth}
            \centerline{\includegraphics[width=\linewidth]{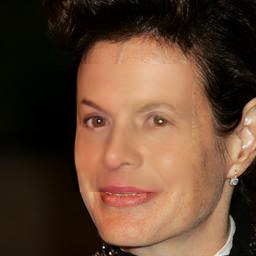}}
            \centerline{(h) PEN}
        \end{minipage}
        \begin{minipage}[t]{0.10\textwidth}
            \centerline{\includegraphics[width=\linewidth]{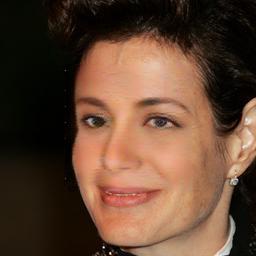}}
            \centerline{(I) DEGNet}
        \end{minipage}

    \end{center}
    \caption{Qualitative results of different inpainting methods on the side face.}
    \label{fig_qualitative_side}
    %\vskip -0.2in
\end{figure*}

\clearpage
}

\textbf{(2) Face inpainting methods:} Besides comparing to the general image inpainting methods, we also compare to face inpainting methods. GFCNet~\cite{li2017generative}, and CollaGAN~\cite{liao2018face}. As there is no public code available for both methods, we report the best performance in their paper.  Aiming at the fair comparison, our experiments follow their experiment setting and use the same dataset with the same training and testing split.

\begin{table}[t]
%\vspace{-0.4cm}
%\setlength{\abovecaptionskip}{-0.01cm}
%\setlength{\belowcaptionskip}{-80pt}
\caption{Qualitative face completion comparison of different models with different settings and varying numbers of tasks: left eye (O1), right eye (O2), upper face (O3) left face (O4), right face (O5), and lower face (O6). The numbers are SSIM/PSNR, the higher the better.}
\label{tb_CelebAHQ-O1-O6}
\centering
\begin{tabular}{c|r|r|r|r}
\hline
Method & CE~\cite{pathak2016context} & GFCNet~\cite{li2017generative} & CollarGAN~\cite{liao2018face} &DEGNet \\
\hline
O1 & 0.905/26.74 & 0.789/19.7 & 0.924/27.76 & \textbf{0.994/39.69} \\
O2 & 0.906/27.01 & 0.784/19.5 & 0.926/27.97 & \textbf{0.994/40.09} \\
O3 & 0.938/27.90 & 0.759/18.8 & 0.952/28.79 & \textbf{0.986/35.50} \\
O4 & 0.958/30.37 & 0.824/20.0 & \textbf{0.972}/31.44 & 0.966/\textbf{31.96} \\
O5 & 0.960/30.65 & 0.826/19.8 & 0.972/31.5 & \textbf{0.975/33.97} \\
O6 & 0.90/27.11 & 0.841/20.2 & 0.917/27.81 & \textbf{0.990/39.14} \\
\hline
    \end{tabular}
%    \vspace{-0.6cm}
\end{table}

\begin{table}[t]
%\vspace{-0.4cm}
\setlength{\abovecaptionskip}{-0.01cm}
\caption{Comparison of the proposed DEGNet with the other advanced methods on the side face.}
\label{tb_CelebAHQ_qualitative_side}
    \centering
    \begin{tabular}{c|r|r}
        \hline
Method &  PSNR & SSIM \\
\hline
PM~\cite{barnes2009patchmatch} &24.44 &0.873  \\
CE~\cite{pathak2016context} &25.35  &0.878 \\
GLCIC~\cite{pathak2016context} &25.06 &0.864 \\
CA \cite{yu2018generative} & 22.4 & 0.844  \\
PICNet \cite{zheng2019pluralistic} & 24.87 & 0.867  \\
PEN \cite{yan2019PENnet} & 25.24 & 0.872  \\
DEGNet (Our)& \textbf{25.36} & \textbf{0.886}\\
\hline
    \end{tabular}
%    \vspace{-0.6cm}
\end{table}

\begin{figure}[t]
    %\vskip 0.2in
    \begin{center}
  %  \scalebox{0.8}{
        \begin{minipage}[t]{0.07\textwidth}
            \includegraphics[width=\linewidth]{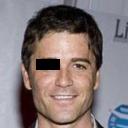}
        \end{minipage}
        \begin{minipage}[t]{0.07\textwidth}
            \includegraphics[width=\linewidth]{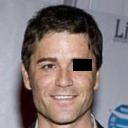}
        \end{minipage}
        \begin{minipage}[t]{0.07\textwidth}
            \includegraphics[width=\linewidth]{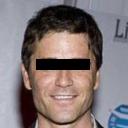}
        \end{minipage}
        \begin{minipage}[t]{0.07\textwidth}
            \includegraphics[width=\linewidth]{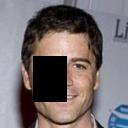}
        \end{minipage}
        \begin{minipage}[t]{0.07\textwidth}
            \includegraphics[width=\linewidth]{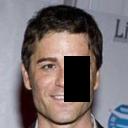}
        \end{minipage}
        \begin{minipage}[t]{0.07\textwidth}
            \includegraphics[width=\linewidth]{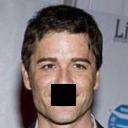}
        \end{minipage}

        \begin{minipage}[t]{0.07\textwidth}
            \includegraphics[width=\linewidth]{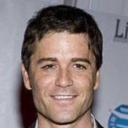}
                                    %\centerline{(a) O1}
        \end{minipage}
        \begin{minipage}[t]{0.07\textwidth}
            \includegraphics[width=\linewidth]{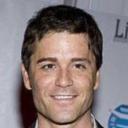}
                                   % \centerline{(b) O2}
        \end{minipage}
        \begin{minipage}[t]{0.07\textwidth}
            \includegraphics[width=\linewidth]{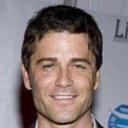}
                                   % \centerline{(c) O3}
        \end{minipage}
        \begin{minipage}[t]{0.07\textwidth}
            \includegraphics[width=\linewidth]{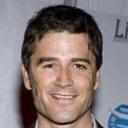}
                                   % \centerline{(d) O4}
        \end{minipage}
        \begin{minipage}[t]{0.07\textwidth}
            \centerline{\includegraphics[width=\linewidth]{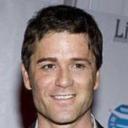}}
            %\centerline{(e) O5}
        \end{minipage}
        \begin{minipage}[t]{0.07\textwidth}
            \centerline{\includegraphics[width=\linewidth]{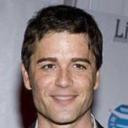}}
           % \centerline{(f) O6}
        \end{minipage}

        \begin{minipage}[t]{0.07\textwidth}
            \includegraphics[width=\linewidth]{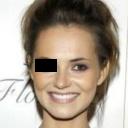}
        \end{minipage}
        \begin{minipage}[t]{0.07\textwidth}
            \includegraphics[width=\linewidth]{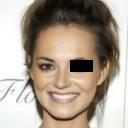}
        \end{minipage}
        \begin{minipage}[t]{0.07\textwidth}
            \includegraphics[width=\linewidth]{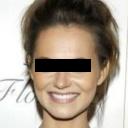}
        \end{minipage}
        \begin{minipage}[t]{0.07\textwidth}
            \includegraphics[width=\linewidth]{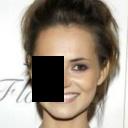}
        \end{minipage}
        \begin{minipage}[t]{0.07\textwidth}
            \includegraphics[width=\linewidth]{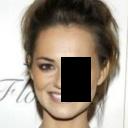}
        \end{minipage}
        \begin{minipage}[t]{0.07\textwidth}
            \includegraphics[width=\linewidth]{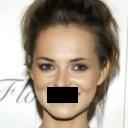}
        \end{minipage}

        \begin{minipage}[t]{0.07\textwidth}
            \includegraphics[width=\linewidth]{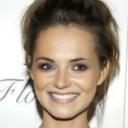}
                                    %\centerline{(a) O1}
        \end{minipage}
        \begin{minipage}[t]{0.07\textwidth}
            \includegraphics[width=\linewidth]{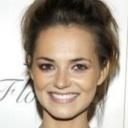}
                                   % \centerline{(b) O2}
        \end{minipage}
        \begin{minipage}[t]{0.07\textwidth}
            \includegraphics[width=\linewidth]{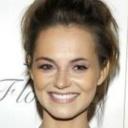}
                                   % \centerline{(c) O3}
        \end{minipage}
        \begin{minipage}[t]{0.07\textwidth}
            \includegraphics[width=\linewidth]{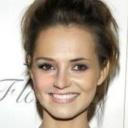}
                                   % \centerline{(d) O4}
        \end{minipage}
        \begin{minipage}[t]{0.07\textwidth}
            \centerline{\includegraphics[width=\linewidth]{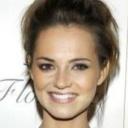}}
            %\centerline{(e) O5}
        \end{minipage}
        \begin{minipage}[t]{0.07\textwidth}
            \centerline{\includegraphics[width=\linewidth]{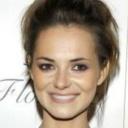}}
           % \centerline{(f) O6}
        \end{minipage}

        \begin{minipage}[t]{0.07\textwidth}
            \includegraphics[width=\linewidth]{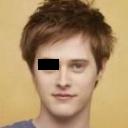}
        \end{minipage}
        \begin{minipage}[t]{0.07\textwidth}
            \includegraphics[width=\linewidth]{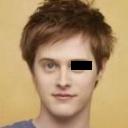}
        \end{minipage}
        \begin{minipage}[t]{0.07\textwidth}
            \includegraphics[width=\linewidth]{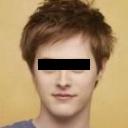}
        \end{minipage}
        \begin{minipage}[t]{0.07\textwidth}
            \includegraphics[width=\linewidth]{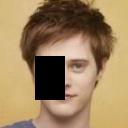}
        \end{minipage}
        \begin{minipage}[t]{0.07\textwidth}
            \includegraphics[width=\linewidth]{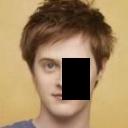}
        \end{minipage}
        \begin{minipage}[t]{0.07\textwidth}
            \includegraphics[width=\linewidth]{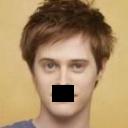}
        \end{minipage}

        \begin{minipage}[t]{0.07\textwidth}
            \includegraphics[width=\linewidth]{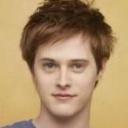}
                                    %\centerline{(a) O1}
        \end{minipage}
        \begin{minipage}[t]{0.07\textwidth}
            \includegraphics[width=\linewidth]{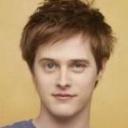}
                                   % \centerline{(b) O2}
        \end{minipage}
        \begin{minipage}[t]{0.07\textwidth}
            \includegraphics[width=\linewidth]{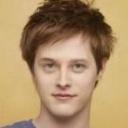}
                                   % \centerline{(c) O3}
        \end{minipage}
        \begin{minipage}[t]{0.07\textwidth}
            \includegraphics[width=\linewidth]{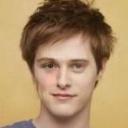}
                                   % \centerline{(d) O4}
        \end{minipage}
        \begin{minipage}[t]{0.07\textwidth}
            \centerline{\includegraphics[width=\linewidth]{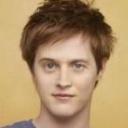}}
            %\centerline{(e) O5}
        \end{minipage}
        \begin{minipage}[t]{0.07\textwidth}
            \centerline{\includegraphics[width=\linewidth]{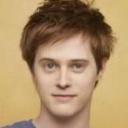}}
           % \centerline{(f) O6}
        \end{minipage}

        \begin{minipage}[t]{0.07\textwidth}
            \includegraphics[width=\linewidth]{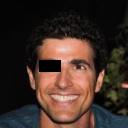}
        \end{minipage}
        \begin{minipage}[t]{0.07\textwidth}
            \includegraphics[width=\linewidth]{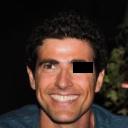}
        \end{minipage}
        \begin{minipage}[t]{0.07\textwidth}
            \includegraphics[width=\linewidth]{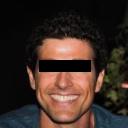}
        \end{minipage}
        \begin{minipage}[t]{0.07\textwidth}
            \includegraphics[width=\linewidth]{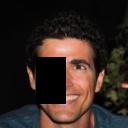}
        \end{minipage}
        \begin{minipage}[t]{0.07\textwidth}
            \includegraphics[width=\linewidth]{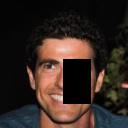}
        \end{minipage}
        \begin{minipage}[t]{0.07\textwidth}
            \includegraphics[width=\linewidth]{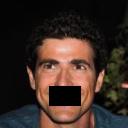}
        \end{minipage}

        \begin{minipage}[t]{0.07\textwidth}
            \includegraphics[width=\linewidth]{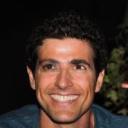}
                                    %\centerline{(a) O1}
        \end{minipage}
        \begin{minipage}[t]{0.07\textwidth}
            \includegraphics[width=\linewidth]{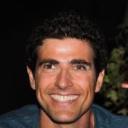}
                                   % \centerline{(b) O2}
        \end{minipage}
        \begin{minipage}[t]{0.07\textwidth}
            \includegraphics[width=\linewidth]{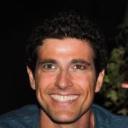}
                                   % \centerline{(c) O3}
        \end{minipage}
        \begin{minipage}[t]{0.07\textwidth}
            \includegraphics[width=\linewidth]{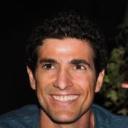}
                                   % \centerline{(d) O4}
        \end{minipage}
        \begin{minipage}[t]{0.07\textwidth}
            \centerline{\includegraphics[width=\linewidth]{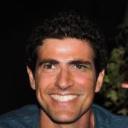}}
            %\centerline{(e) O5}
        \end{minipage}
        \begin{minipage}[t]{0.07\textwidth}
            \centerline{\includegraphics[width=\linewidth]{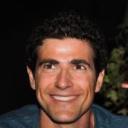}}
           % \centerline{(f) O6}
        \end{minipage}

        \begin{minipage}[t]{0.07\textwidth}
            \includegraphics[width=\linewidth]{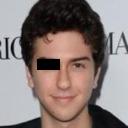}
        \end{minipage}
        \begin{minipage}[t]{0.07\textwidth}
            \includegraphics[width=\linewidth]{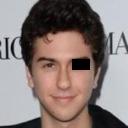}
        \end{minipage}
        \begin{minipage}[t]{0.07\textwidth}
            \includegraphics[width=\linewidth]{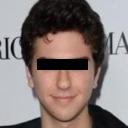}
        \end{minipage}
        \begin{minipage}[t]{0.07\textwidth}
            \includegraphics[width=\linewidth]{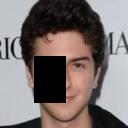}
        \end{minipage}
        \begin{minipage}[t]{0.07\textwidth}
            \includegraphics[width=\linewidth]{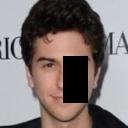}
        \end{minipage}
        \begin{minipage}[t]{0.07\textwidth}
            \includegraphics[width=\linewidth]{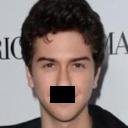}
        \end{minipage}

        \begin{minipage}[t]{0.07\textwidth}
            \includegraphics[width=\linewidth]{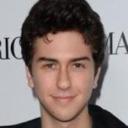}
                                    %\centerline{(a) O1}
        \end{minipage}
        \begin{minipage}[t]{0.07\textwidth}
            \includegraphics[width=\linewidth]{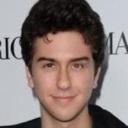}
                                   % \centerline{(b) O2}
        \end{minipage}
        \begin{minipage}[t]{0.07\textwidth}
            \includegraphics[width=\linewidth]{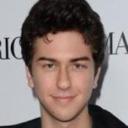}
                                   % \centerline{(c) O3}
        \end{minipage}
        \begin{minipage}[t]{0.07\textwidth}
            \includegraphics[width=\linewidth]{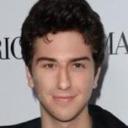}
                                   % \centerline{(d) O4}
        \end{minipage}
        \begin{minipage}[t]{0.07\textwidth}
            \centerline{\includegraphics[width=\linewidth]{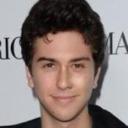}}
            %\centerline{(e) O5}
        \end{minipage}
        \begin{minipage}[t]{0.07\textwidth}
            \centerline{\includegraphics[width=\linewidth]{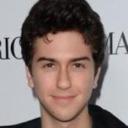}}
           % \centerline{(f) O6}
        \end{minipage}

        \begin{minipage}[t]{0.07\textwidth}
            \includegraphics[width=\linewidth]{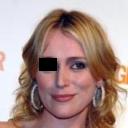}
        \end{minipage}
        \begin{minipage}[t]{0.07\textwidth}
            \includegraphics[width=\linewidth]{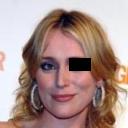}
        \end{minipage}
        \begin{minipage}[t]{0.07\textwidth}
            \includegraphics[width=\linewidth]{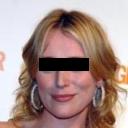}
        \end{minipage}
        \begin{minipage}[t]{0.07\textwidth}
            \includegraphics[width=\linewidth]{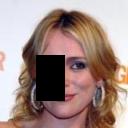}
        \end{minipage}
        \begin{minipage}[t]{0.07\textwidth}
            \includegraphics[width=\linewidth]{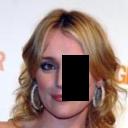}
        \end{minipage}
        \begin{minipage}[t]{0.07\textwidth}
            \includegraphics[width=\linewidth]{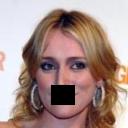}
        \end{minipage}

        \begin{minipage}[t]{0.07\textwidth}
            \includegraphics[width=\linewidth]{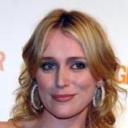}
                                    %\centerline{(a) O1}
        \end{minipage}
        \begin{minipage}[t]{0.07\textwidth}
            \includegraphics[width=\linewidth]{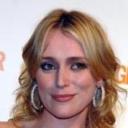}
                                   % \centerline{(b) O2}
        \end{minipage}
        \begin{minipage}[t]{0.07\textwidth}
            \includegraphics[width=\linewidth]{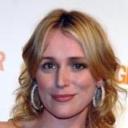}
                                   % \centerline{(c) O3}
        \end{minipage}
        \begin{minipage}[t]{0.07\textwidth}
            \includegraphics[width=\linewidth]{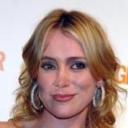}
                                   % \centerline{(d) O4}
        \end{minipage}
        \begin{minipage}[t]{0.07\textwidth}
            \centerline{\includegraphics[width=\linewidth]{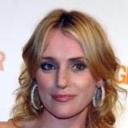}}
            %\centerline{(e) O5}
        \end{minipage}
        \begin{minipage}[t]{0.07\textwidth}
            \centerline{\includegraphics[width=\linewidth]{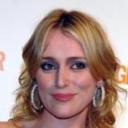}}
           % \centerline{(f) O6}
        \end{minipage}

        \begin{minipage}[t]{0.07\textwidth}
            \includegraphics[width=\linewidth]{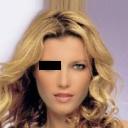}
        \end{minipage}
        \begin{minipage}[t]{0.07\textwidth}
            \includegraphics[width=\linewidth]{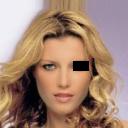}
        \end{minipage}
        \begin{minipage}[t]{0.07\textwidth}
            \includegraphics[width=\linewidth]{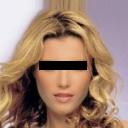}
        \end{minipage}
        \begin{minipage}[t]{0.07\textwidth}
            \includegraphics[width=\linewidth]{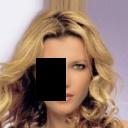}
        \end{minipage}
        \begin{minipage}[t]{0.07\textwidth}
            \includegraphics[width=\linewidth]{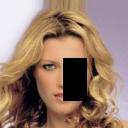}
        \end{minipage}
        \begin{minipage}[t]{0.07\textwidth}
            \includegraphics[width=\linewidth]{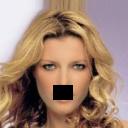}
        \end{minipage}

        \begin{minipage}[t]{0.07\textwidth}
            \includegraphics[width=\linewidth]{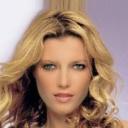}
                                    %\centerline{(a) O1}
        \end{minipage}
        \begin{minipage}[t]{0.07\textwidth}
            \includegraphics[width=\linewidth]{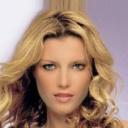}
                                   % \centerline{(b) O2}
        \end{minipage}
        \begin{minipage}[t]{0.07\textwidth}
            \includegraphics[width=\linewidth]{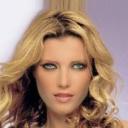}
                                   % \centerline{(c) O3}
        \end{minipage}
        \begin{minipage}[t]{0.07\textwidth}
            \includegraphics[width=\linewidth]{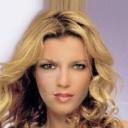}
                                   % \centerline{(d) O4}
        \end{minipage}
        \begin{minipage}[t]{0.07\textwidth}
            \centerline{\includegraphics[width=\linewidth]{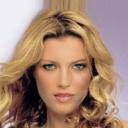}}
            %\centerline{(e) O5}
        \end{minipage}
        \begin{minipage}[t]{0.07\textwidth}
            \centerline{\includegraphics[width=\linewidth]{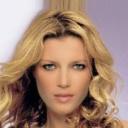}}
           % \centerline{(f) O6}
        \end{minipage}

        \begin{minipage}[t]{0.07\textwidth}
            \includegraphics[width=\linewidth]{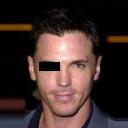}
        \end{minipage}
        \begin{minipage}[t]{0.07\textwidth}
            \includegraphics[width=\linewidth]{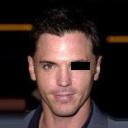}
        \end{minipage}
        \begin{minipage}[t]{0.07\textwidth}
            \includegraphics[width=\linewidth]{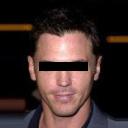}
        \end{minipage}
        \begin{minipage}[t]{0.07\textwidth}
            \includegraphics[width=\linewidth]{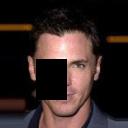}
        \end{minipage}
        \begin{minipage}[t]{0.07\textwidth}
            \includegraphics[width=\linewidth]{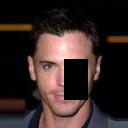}
        \end{minipage}
        \begin{minipage}[t]{0.07\textwidth}
            \includegraphics[width=\linewidth]{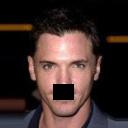}
        \end{minipage}

        \begin{minipage}[t]{0.07\textwidth}
            \includegraphics[width=\linewidth]{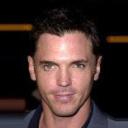}
                                    %\centerline{(a) O1}
        \end{minipage}
        \begin{minipage}[t]{0.07\textwidth}
            \includegraphics[width=\linewidth]{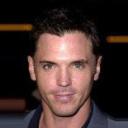}
                                   % \centerline{(b) O2}
        \end{minipage}
        \begin{minipage}[t]{0.07\textwidth}
            \includegraphics[width=\linewidth]{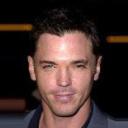}
                                   % \centerline{(c) O3}
        \end{minipage}
        \begin{minipage}[t]{0.07\textwidth}
            \includegraphics[width=\linewidth]{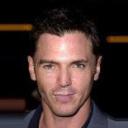}
                                   % \centerline{(d) O4}
        \end{minipage}
        \begin{minipage}[t]{0.07\textwidth}
            \centerline{\includegraphics[width=\linewidth]{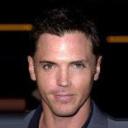}}
            %\centerline{(e) O5}
        \end{minipage}
        \begin{minipage}[t]{0.07\textwidth}
            \centerline{\includegraphics[width=\linewidth]{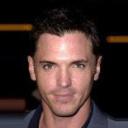}}
           % \centerline{(f) O6}
        \end{minipage}

        \begin{minipage}[t]{0.07\textwidth}
            \includegraphics[width=\linewidth]{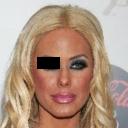}
        \end{minipage}
        \begin{minipage}[t]{0.07\textwidth}
            \includegraphics[width=\linewidth]{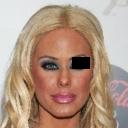}
        \end{minipage}
        \begin{minipage}[t]{0.07\textwidth}
            \includegraphics[width=\linewidth]{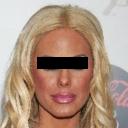}
        \end{minipage}
        \begin{minipage}[t]{0.07\textwidth}
            \includegraphics[width=\linewidth]{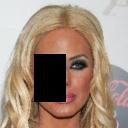}
        \end{minipage}
        \begin{minipage}[t]{0.07\textwidth}
            \includegraphics[width=\linewidth]{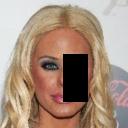}
        \end{minipage}
        \begin{minipage}[t]{0.07\textwidth}
            \includegraphics[width=\linewidth]{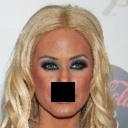}
        \end{minipage}

        \begin{minipage}[t]{0.07\textwidth}
            \includegraphics[width=\linewidth]{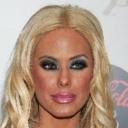}
                                    \centerline{(a) O1}
        \end{minipage}
        \begin{minipage}[t]{0.07\textwidth}
            \includegraphics[width=\linewidth]{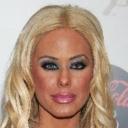}
                                    \centerline{(b) O2}
        \end{minipage}
        \begin{minipage}[t]{0.07\textwidth}
            \includegraphics[width=\linewidth]{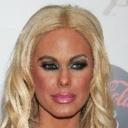}
                                    \centerline{(c) O3}
        \end{minipage}
        \begin{minipage}[t]{0.07\textwidth}
            \includegraphics[width=\linewidth]{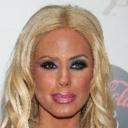}
                                    \centerline{(d) O4}
        \end{minipage}
        \begin{minipage}[t]{0.07\textwidth}
            \centerline{\includegraphics[width=\linewidth]{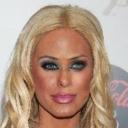}}
            \centerline{(e) O5}
        \end{minipage}
        \begin{minipage}[t]{0.07\textwidth}
            \centerline{\includegraphics[width=\linewidth]{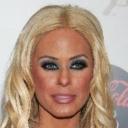}}
            \centerline{(f) O6}
        \end{minipage}
 %       }
    \end{center}
    \caption{Qualitative face completion comparison of our methods with different masks.}
    \label{fig_O1_O6_qualititive}
    \vskip -0.2in
\end{figure}

{As shown in Table~\ref{tb_CelebAHQ-O1-O6}, our method, in general, achieves better performance over all other methods in terms of SSIM and PSNR. It is easy to find that the values of PSNR and SSIM in our methods are significantly higher than those of GFCNet with CollaGAN except O4.}

 As shown in Fig.~\ref{fig_O1_O6_qualititive}, we completed all kinds of part cropped face inpainting. We produced a clear and natural result. Because without code and visual features in their paper, we only show our features.

\begin{figure}[t]
    %\vskip 0.2in
    \begin{center}
        \begin{minipage}[t]{0.09\textwidth}
            \includegraphics[width=\linewidth]{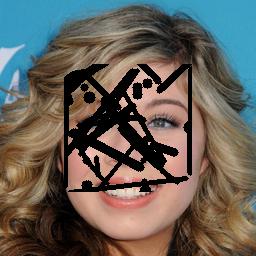}
        \end{minipage}
        \begin{minipage}[t]{0.09\textwidth}
            \includegraphics[width=\linewidth]{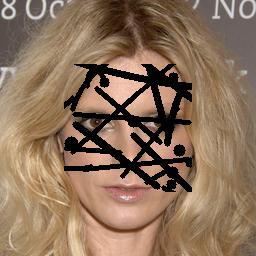}
        \end{minipage}
        \begin{minipage}[t]{0.09\textwidth}
            \includegraphics[width=\linewidth]{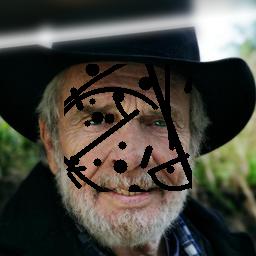}
        \end{minipage}
        \begin{minipage}[t]{0.09\textwidth}
            \includegraphics[width=\linewidth]{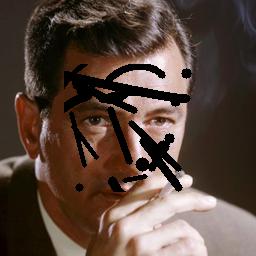}
        \end{minipage}
        \begin{minipage}[t]{0.09\textwidth}
            \includegraphics[width=\linewidth]{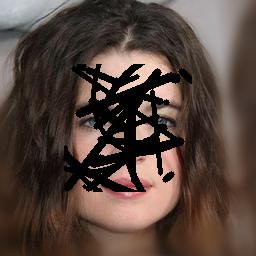}
        \end{minipage}

        \begin{minipage}[t]{0.09\textwidth}
            \includegraphics[width=\linewidth]{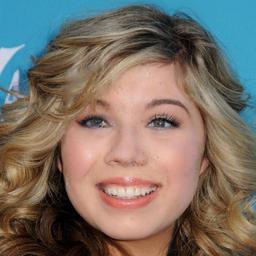}
        \end{minipage}
        \begin{minipage}[t]{0.09\textwidth}
            \includegraphics[width=\linewidth]{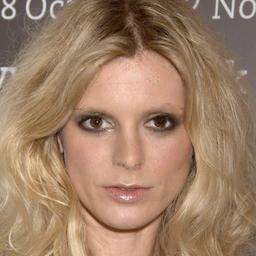}
        \end{minipage}
        \begin{minipage}[t]{0.09\textwidth}
            \includegraphics[width=\linewidth]{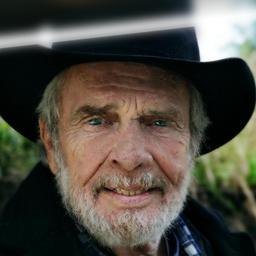}
        \end{minipage}
        \begin{minipage}[t]{0.09\textwidth}
            \includegraphics[width=\linewidth]{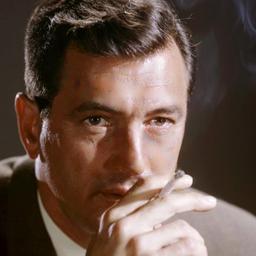}
        \end{minipage}
        \begin{minipage}[t]{0.09\textwidth}
            \includegraphics[width=\linewidth]{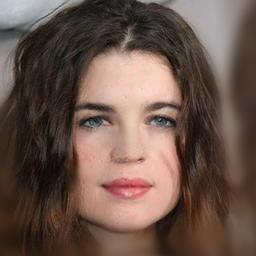}
        \end{minipage}

        \begin{minipage}[t]{0.09\textwidth}
            \includegraphics[width=\linewidth]{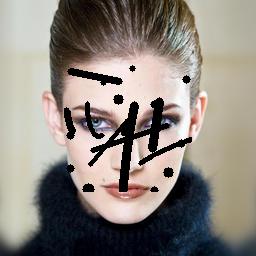}
        \end{minipage}
        \begin{minipage}[t]{0.09\textwidth}
            \includegraphics[width=\linewidth]{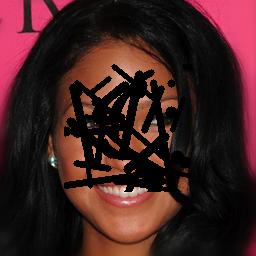}
        \end{minipage}
        \begin{minipage}[t]{0.09\textwidth}
            \includegraphics[width=\linewidth]{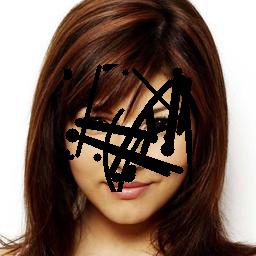}
        \end{minipage}
        \begin{minipage}[t]{0.09\textwidth}
            \includegraphics[width=\linewidth]{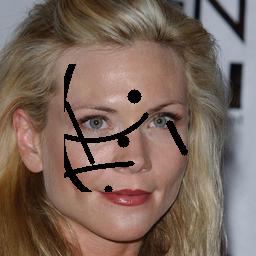}
        \end{minipage}
        \begin{minipage}[t]{0.09\textwidth}
            \includegraphics[width=\linewidth]{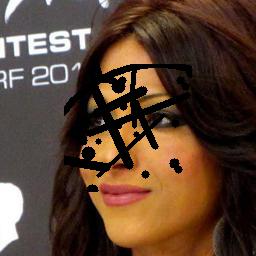}
        \end{minipage}

        \begin{minipage}[t]{0.09\textwidth}
            \includegraphics[width=\linewidth]{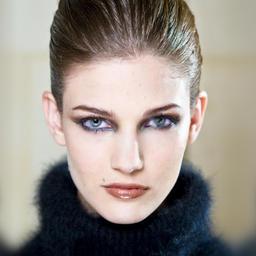}
        \end{minipage}
        \begin{minipage}[t]{0.09\textwidth}
            \includegraphics[width=\linewidth]{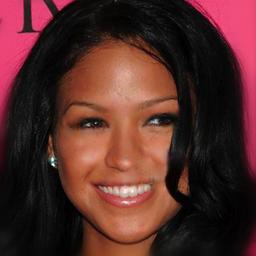}
        \end{minipage}
        \begin{minipage}[t]{0.09\textwidth}
            \includegraphics[width=\linewidth]{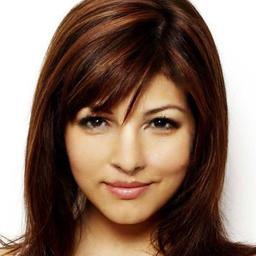}
        \end{minipage}
        \begin{minipage}[t]{0.09\textwidth}
            \includegraphics[width=\linewidth]{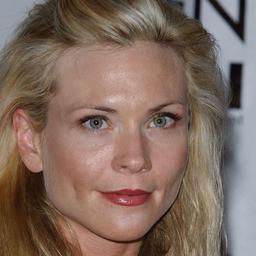}
        \end{minipage}
        \begin{minipage}[t]{0.09\textwidth}
            \includegraphics[width=\linewidth]{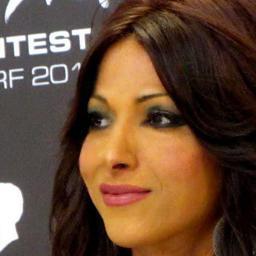}
        \end{minipage}

        \begin{minipage}[t]{0.09\textwidth}
            \includegraphics[width=\linewidth]{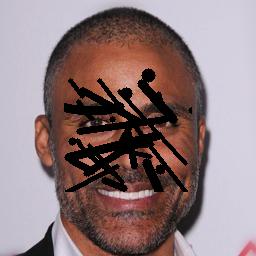}
        \end{minipage}
        \begin{minipage}[t]{0.09\textwidth}
            \includegraphics[width=\linewidth]{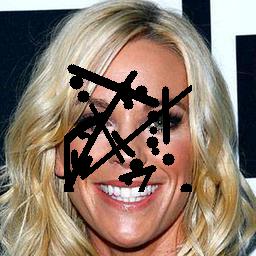}
        \end{minipage}
        \begin{minipage}[t]{0.09\textwidth}
            \includegraphics[width=\linewidth]{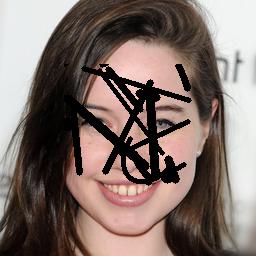}
        \end{minipage}
        \begin{minipage}[t]{0.09\textwidth}
            \includegraphics[width=\linewidth]{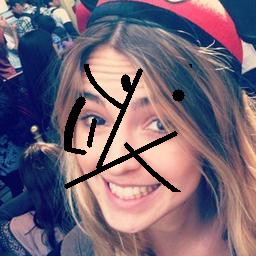}
        \end{minipage}
        \begin{minipage}[t]{0.09\textwidth}
            \includegraphics[width=\linewidth]{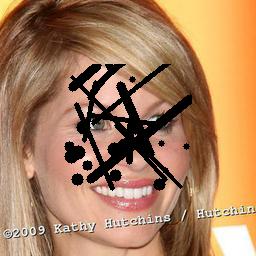}
        \end{minipage}

        \begin{minipage}[t]{0.09\textwidth}
            \includegraphics[width=\linewidth]{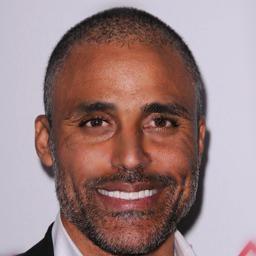}
        \end{minipage}
        \begin{minipage}[t]{0.09\textwidth}
            \includegraphics[width=\linewidth]{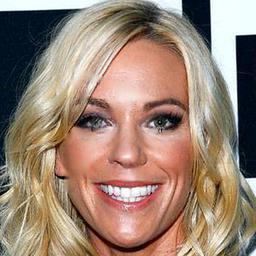}
        \end{minipage}
        \begin{minipage}[t]{0.09\textwidth}
            \includegraphics[width=\linewidth]{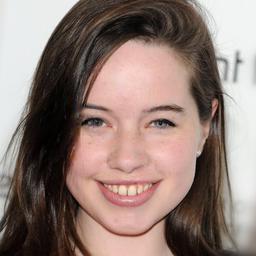}
        \end{minipage}
        \begin{minipage}[t]{0.09\textwidth}
            \includegraphics[width=\linewidth]{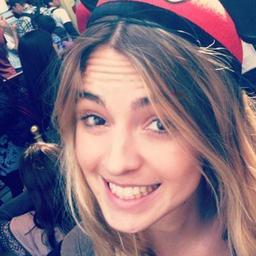}
        \end{minipage}
        \begin{minipage}[t]{0.09\textwidth}
            \includegraphics[width=\linewidth]{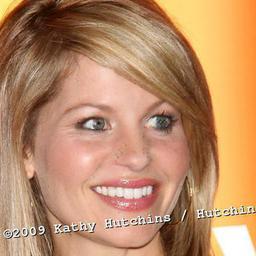}
        \end{minipage}

        \begin{minipage}[t]{0.09\textwidth}
            \includegraphics[width=\linewidth]{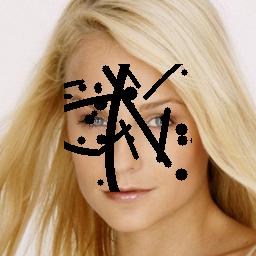}
        \end{minipage}
        \begin{minipage}[t]{0.09\textwidth}
            \includegraphics[width=\linewidth]{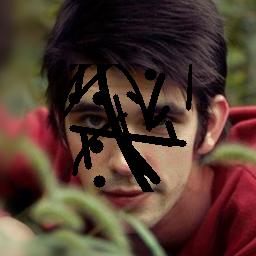}
        \end{minipage}
        \begin{minipage}[t]{0.09\textwidth}
            \includegraphics[width=\linewidth]{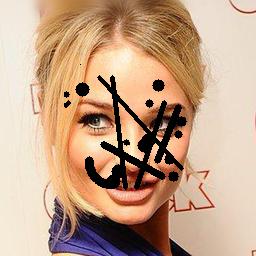}
        \end{minipage}
        \begin{minipage}[t]{0.09\textwidth}
            \includegraphics[width=\linewidth]{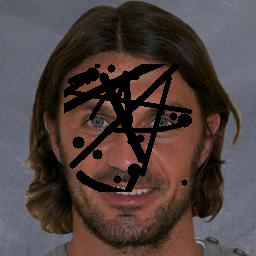}
        \end{minipage}
        \begin{minipage}[t]{0.09\textwidth}
            \includegraphics[width=\linewidth]{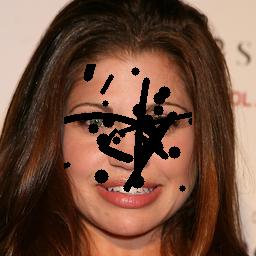}
        \end{minipage}

        \begin{minipage}[t]{0.09\textwidth}
            \includegraphics[width=\linewidth]{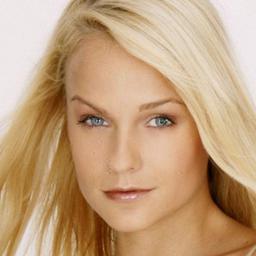}
        \end{minipage}
        \begin{minipage}[t]{0.09\textwidth}
            \includegraphics[width=\linewidth]{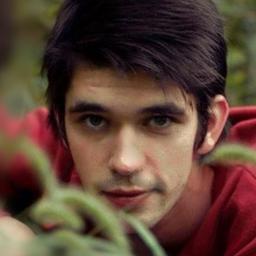}
        \end{minipage}
        \begin{minipage}[t]{0.09\textwidth}
            \includegraphics[width=\linewidth]{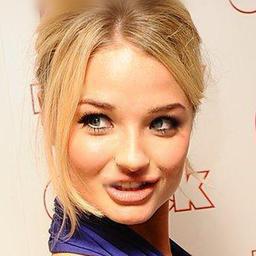}
        \end{minipage}
        \begin{minipage}[t]{0.09\textwidth}
            \includegraphics[width=\linewidth]{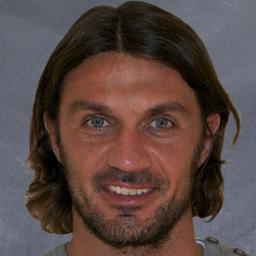}
        \end{minipage}
        \begin{minipage}[t]{0.09\textwidth}
            \includegraphics[width=\linewidth]{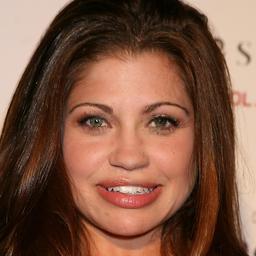}
        \end{minipage}

        \begin{minipage}[t]{0.09\textwidth}
            \includegraphics[width=\linewidth]{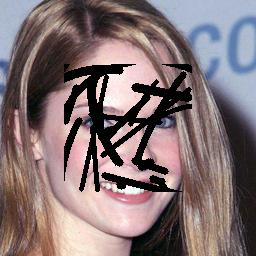}
        \end{minipage}
        \begin{minipage}[t]{0.09\textwidth}
            \includegraphics[width=\linewidth]{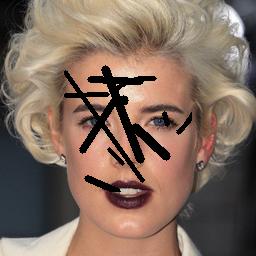}
        \end{minipage}
        \begin{minipage}[t]{0.09\textwidth}
            \includegraphics[width=\linewidth]{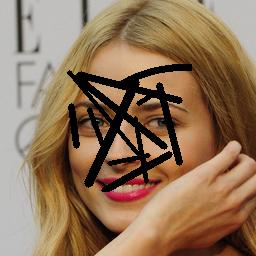}
        \end{minipage}
        \begin{minipage}[t]{0.09\textwidth}
            \includegraphics[width=\linewidth]{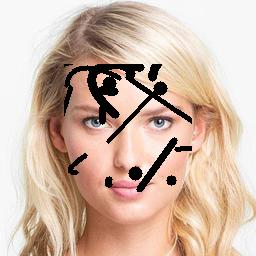}
        \end{minipage}
        \begin{minipage}[t]{0.09\textwidth}
            \includegraphics[width=\linewidth]{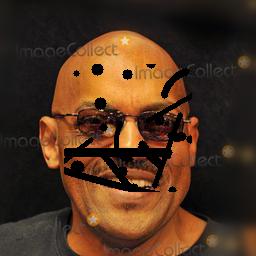}
        \end{minipage}

        \begin{minipage}[t]{0.09\textwidth}
            \includegraphics[width=\linewidth]{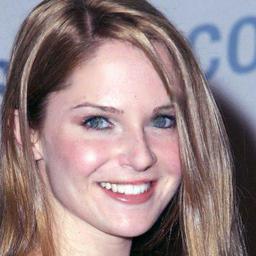}
        \end{minipage}
        \begin{minipage}[t]{0.09\textwidth}
            \includegraphics[width=\linewidth]{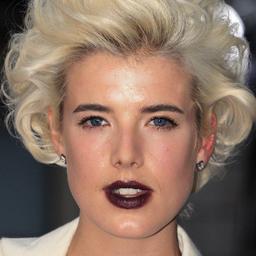}
        \end{minipage}
        \begin{minipage}[t]{0.09\textwidth}
            \includegraphics[width=\linewidth]{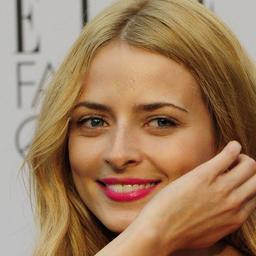}
        \end{minipage}
        \begin{minipage}[t]{0.09\textwidth}
            \includegraphics[width=\linewidth]{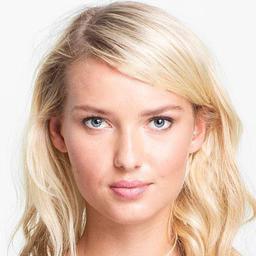}
        \end{minipage}
        \begin{minipage}[t]{0.09\textwidth}
            \includegraphics[width=\linewidth]{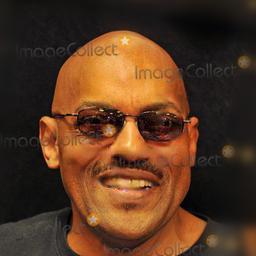}
        \end{minipage}
        \end{center}
    \caption{Our method is used to complete the effect of irregular missing area.}
    \label{irr_fig}
    %\vskip -0.2in
\end{figure}

In addition, our method is also applicable to the completion of irregular missing areas. As shown in Fig.~\ref{irr_fig}, our method also achieves good visual quality for irregular completion.

\subsection{Evaluation of Side Face Impainting}

Besides frontal face inpainting, we further evaluate face inpainting performance on the side face. Different from the frontal face, the cropped side face contains more missing information. In terms of side face, the facial features information is difficult to learn than the frontal face. So it is hard to complete face inpainting inside face. Generally, most of the existing methods failed on side face inpainting. To illustrate the problem, Table~\ref{tb_CelebAHQ_qualitative_side} and Fig.~\ref{fig_qualitative_side} shows the quantitative and qualitative comparisons of different methods on side face inpainting. Table~\ref{tb_CelebAHQ_qualitative_side} shows that our method outperforms the state-of-the-art in both PSNR and SSIM on the side face inpainting. From Fig.~\ref{fig_qualitative_side}, we find that our DEGNet method has symmetry faces, such as eyes with the same sizes and colors, while other methods include blurry textures and asymmetry faces.

\subsection{Ablation Study}

\begin{figure}[t]
    %\vskip 0.2in
    \begin{center}
        \begin{minipage}[t]{0.09\textwidth}
            \includegraphics[width=\linewidth]{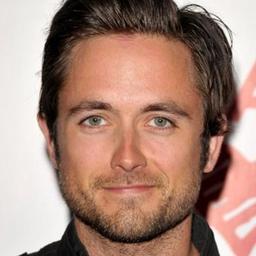}
        \end{minipage}
        \begin{minipage}[t]{0.09\textwidth}
            \includegraphics[width=\linewidth]{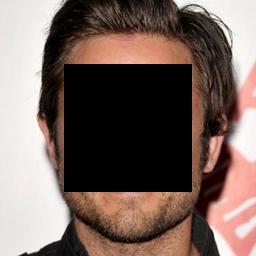}
        \end{minipage}
        \begin{minipage}[t]{0.09\textwidth}
            \includegraphics[width=\linewidth]{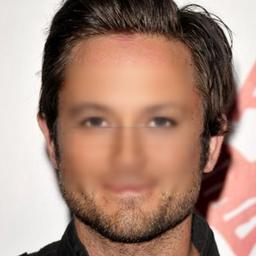}
        \end{minipage}
        \begin{minipage}[t]{0.09\textwidth}
            \includegraphics[width=\linewidth]{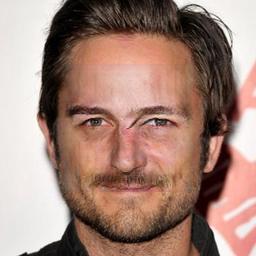}
        \end{minipage}
        \begin{minipage}[t]{0.09\textwidth}
            \includegraphics[width=\linewidth]{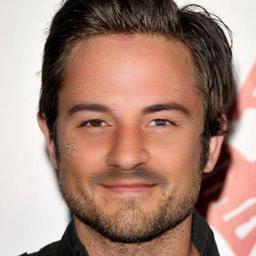}
        \end{minipage}

        \begin{minipage}[t]{0.09\textwidth}
            \includegraphics[width=\linewidth]{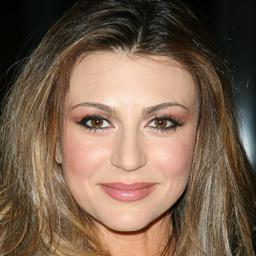}
        \end{minipage}
        \begin{minipage}[t]{0.09\textwidth}
            \includegraphics[width=\linewidth]{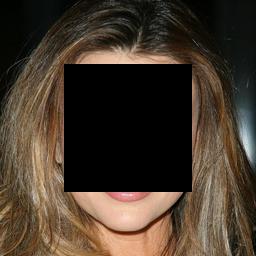}
        \end{minipage}
        \begin{minipage}[t]{0.09\textwidth}
            \includegraphics[width=\linewidth]{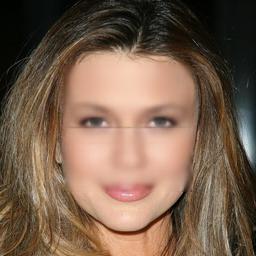}
        \end{minipage}
        \begin{minipage}[t]{0.09\textwidth}
            \includegraphics[width=\linewidth]{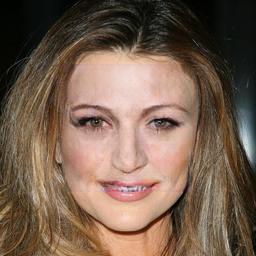}
        \end{minipage}
        \begin{minipage}[t]{0.09\textwidth}
            \includegraphics[width=\linewidth]{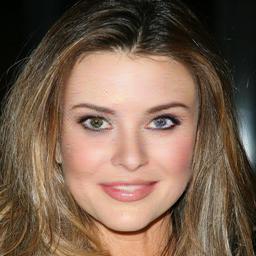}
        \end{minipage}

        \begin{minipage}[t]{0.09\textwidth}
            \includegraphics[width=\linewidth]{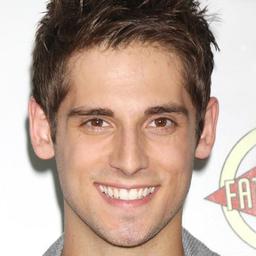}
        \end{minipage}
        \begin{minipage}[t]{0.09\textwidth}
            \includegraphics[width=\linewidth]{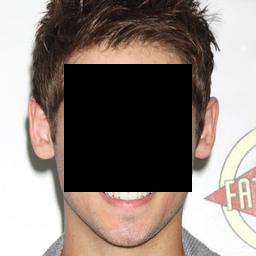}
        \end{minipage}
        \begin{minipage}[t]{0.09\textwidth}
            \includegraphics[width=\linewidth]{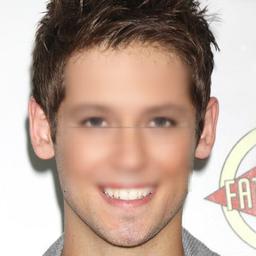}
        \end{minipage}
        \begin{minipage}[t]{0.09\textwidth}
            \includegraphics[width=\linewidth]{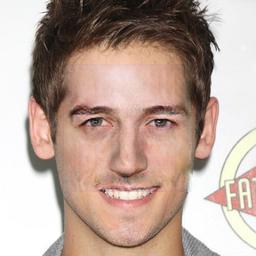}
        \end{minipage}
        \begin{minipage}[t]{0.09\textwidth}
            \includegraphics[width=\linewidth]{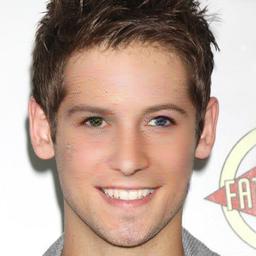}
        \end{minipage}

        \begin{minipage}[t]{0.09\textwidth}
            \includegraphics[width=\linewidth]{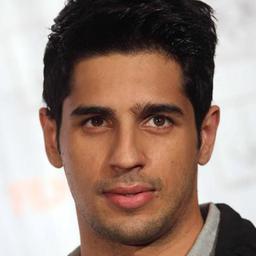}
        \end{minipage}
        \begin{minipage}[t]{0.09\textwidth}
            \includegraphics[width=\linewidth]{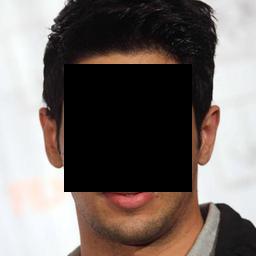}
        \end{minipage}
        \begin{minipage}[t]{0.09\textwidth}
            \includegraphics[width=\linewidth]{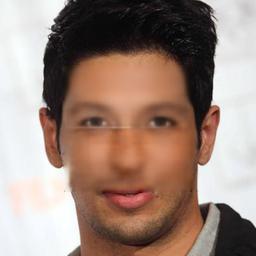}
        \end{minipage}
        \begin{minipage}[t]{0.09\textwidth}
            \includegraphics[width=\linewidth]{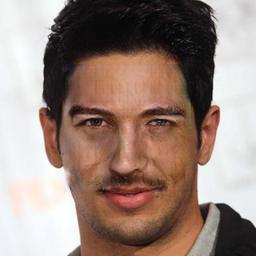}
        \end{minipage}
        \begin{minipage}[t]{0.09\textwidth}
            \includegraphics[width=\linewidth]{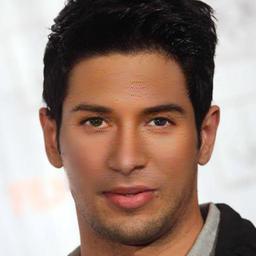}
        \end{minipage}

        \begin{minipage}[t]{0.09\textwidth}
            \includegraphics[width=\linewidth]{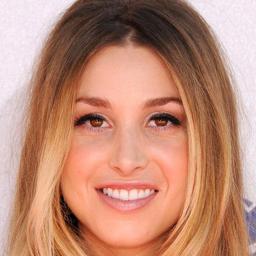}
        \end{minipage}
        \begin{minipage}[t]{0.09\textwidth}
            \includegraphics[width=\linewidth]{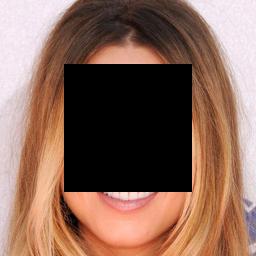}
        \end{minipage}
        \begin{minipage}[t]{0.09\textwidth}
            \includegraphics[width=\linewidth]{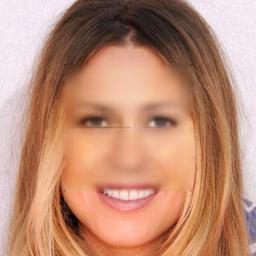}
        \end{minipage}
        \begin{minipage}[t]{0.09\textwidth}
            \includegraphics[width=\linewidth]{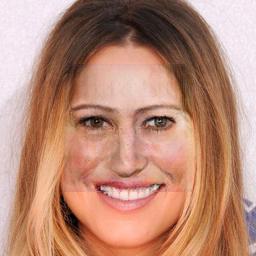}
        \end{minipage}
        \begin{minipage}[t]{0.09\textwidth}
            \includegraphics[width=\linewidth]{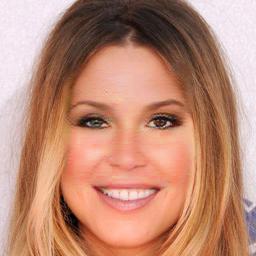}
        \end{minipage}

        \begin{minipage}[t]{0.09\textwidth}
            \includegraphics[width=\linewidth]{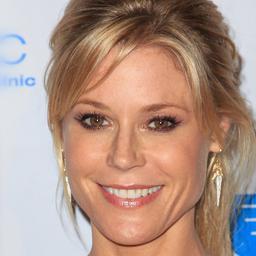}
        \end{minipage}
        \begin{minipage}[t]{0.09\textwidth}
            \includegraphics[width=\linewidth]{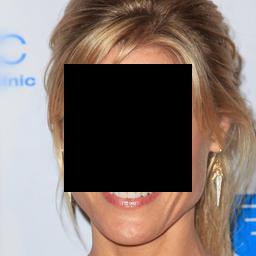}
        \end{minipage}
        \begin{minipage}[t]{0.09\textwidth}
            \includegraphics[width=\linewidth]{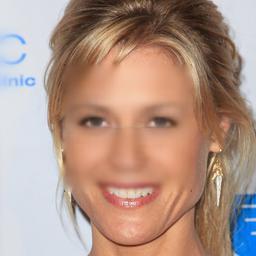}
        \end{minipage}
        \begin{minipage}[t]{0.09\textwidth}
            \includegraphics[width=\linewidth]{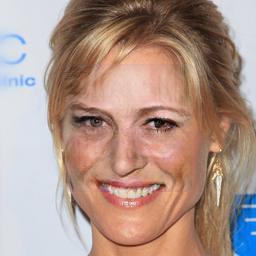}
        \end{minipage}
        \begin{minipage}[t]{0.09\textwidth}
            \includegraphics[width=\linewidth]{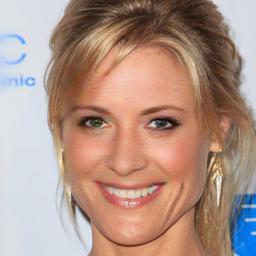}
        \end{minipage}

        \begin{minipage}[t]{0.09\textwidth}
            \includegraphics[width=\linewidth]{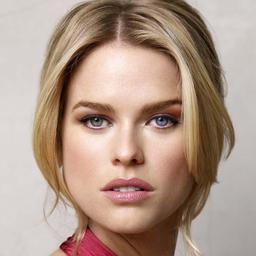}
        \end{minipage}
        \begin{minipage}[t]{0.09\textwidth}
            \includegraphics[width=\linewidth]{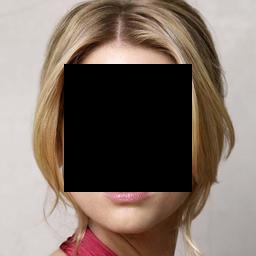}
        \end{minipage}
        \begin{minipage}[t]{0.09\textwidth}
            \includegraphics[width=\linewidth]{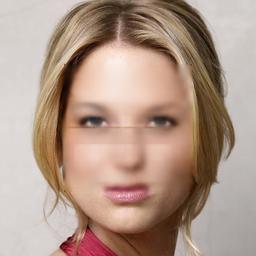}
        \end{minipage}
        \begin{minipage}[t]{0.09\textwidth}
            \includegraphics[width=\linewidth]{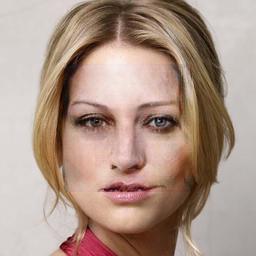}
        \end{minipage}
        \begin{minipage}[t]{0.09\textwidth}
            \includegraphics[width=\linewidth]{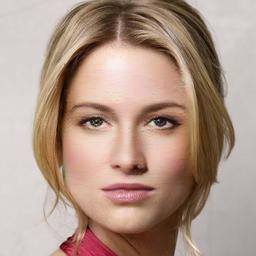}
        \end{minipage}

        \begin{minipage}[t]{0.09\textwidth}
            \includegraphics[width=\linewidth]{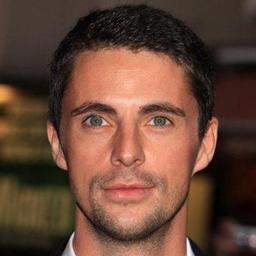}
        \end{minipage}
        \begin{minipage}[t]{0.09\textwidth}
            \includegraphics[width=\linewidth]{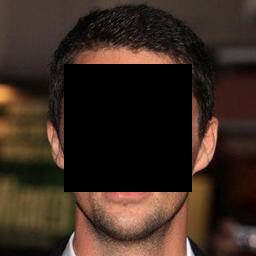}
        \end{minipage}
        \begin{minipage}[t]{0.09\textwidth}
            \includegraphics[width=\linewidth]{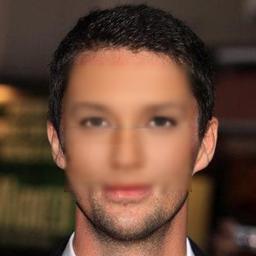}
        \end{minipage}
        \begin{minipage}[t]{0.09\textwidth}
            \includegraphics[width=\linewidth]{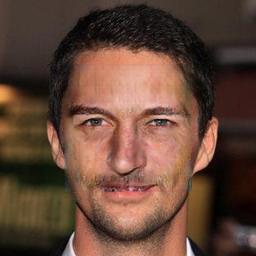}
        \end{minipage}
        \begin{minipage}[t]{0.09\textwidth}
            \includegraphics[width=\linewidth]{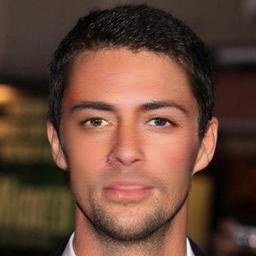}
        \end{minipage}

        \begin{minipage}[t]{0.09\textwidth}
            \includegraphics[width=\linewidth]{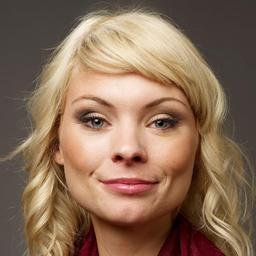}
                                    \centerline{(a) Original}
        \end{minipage}
        \begin{minipage}[t]{0.09\textwidth}
            \includegraphics[width=\linewidth]{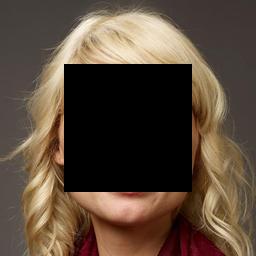}
                                    \centerline{(b) Cropped}
        \end{minipage}
        \begin{minipage}[t]{0.09\textwidth}
            \includegraphics[width=\linewidth]{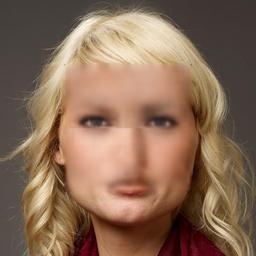}
                                    \centerline{(c) S1}
        \end{minipage}
        \begin{minipage}[t]{0.09\textwidth}
            \includegraphics[width=\linewidth]{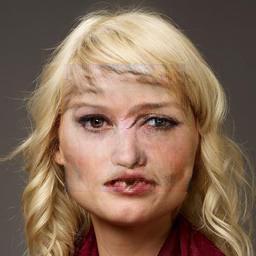}
                                    \centerline{(d) S2}
        \end{minipage}
        \begin{minipage}[t]{0.09\textwidth}
            \centerline{\includegraphics[width=\linewidth]{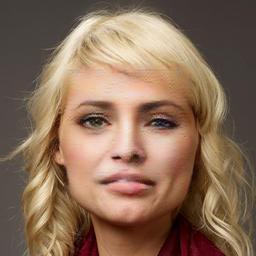}}
            \centerline{(e) S3}
        \end{minipage}
        \end{center}
    \caption{Qualitative face completion comparison of our  methods with different components S1, S2 and S3, and total loss.}
    \label{ablation_fig}
    %\vskip -0.2in
\end{figure}

We further perform experiments to study the effect of the components of our model. We analyze how the different combinations of our components, (S1) reconstruction, (S2) Reconstruction + global discriminator,  (S3) Reconstruction + global discriminator+ patch discriminator, affect our inpainting performance, the results are shown in Table~\ref{Ablation_study} and Fig.~\ref{ablation_fig}).
Based on our backbone, we further impose the latent classifier on the random vector, which results in better PSNR and SSIM.
As an intermediate test, the global discriminator somewhat decreases the SSMI score.
Moreover, based on the former, more other two global and patch discriminators only acting on the missing region would result in more better
results.

And three constraint factors actually can have a positive effect on their performance development.
According to our previous analysis, we know that reconstruction constraints and discriminator can improve the performance of our backbone directly. We also explored different training modes on how to affect its performance. Thus combing with domain embedded generator and discriminator alternative optimization, our DEGNet is proposed to overcome this problem and discuss how to generate high-quality face inpainting images based on this.

\begin{table}[t]
%\vspace{-0.4cm}
\setlength{\abovecaptionskip}{-0.01cm}
    \caption{Quantitative face completion comparison with large regular holes of our methods with different components: rec (S1), rec+global (S2), and rec+global+patch (S3). Higher values are better.}
    \label{Ablation_study}
    \centering
    \begin{tabular}{c|r|r|r}
        \hline
& S1 & S2 & S3\\
\hline
PSNR & 25.590 & 23.518  & \textbf{26.208}\\
SSIM & 0.886 & 0.856  & \textbf{0.895}\\
\hline
\end{tabular}
\end{table}

\section{Conclusion}
We proposed a Domain Embedded Multi-model Generative Adversarial Network for face image inpainting.
Our proposed model improves the face inpainting performance by using the face region information as the guidance information within a Multi-model GAN framework. The prior domain knowledge of face structure information and semantic information are incorprated into the DEGNet to generate a natural and harmonious face. Experimental results demonstrate our method gets better performance than the state-of-the-art face inpainting methods. Furthermore, our method could be easily applied to other image editing tasks.

\bibliographystyle{IEEEtran}
\bibliography{egbib}

\begin{thebibliography}{10}
\providecommand{\url}[1]{#1}
\csname url@samestyle\endcsname
\providecommand{\newblock}{\relax}
\providecommand{\bibinfo}[2]{#2}
\providecommand{\BIBentrySTDinterwordspacing}{\spaceskip=0pt\relax}
\providecommand{\BIBentryALTinterwordstretchfactor}{4}
\providecommand{\BIBentryALTinterwordspacing}{\spaceskip=\fontdimen2\font plus
\BIBentryALTinterwordstretchfactor\fontdimen3\font minus
  \fontdimen4\font\relax}
\providecommand{\BIBforeignlanguage}[2]{{%
\expandafter\ifx\csname l@#1\endcsname\relax
\typeout{** WARNING: IEEEtran.bst: No hyphenation pattern has been}%
\typeout{** loaded for the language `#1'. Using the pattern for}%
\typeout{** the default language instead.}%
\else
\language=\csname l@#1\endcsname
\fi
#2}}
\providecommand{\BIBdecl}{\relax}
\BIBdecl

\bibitem{pathak2016context}
D.~Pathak, P.~Krahenbuhl, J.~Donahue, T.~Darrell, and A.~A. Efros, ``Context
  encoders: Feature learning by inpainting,'' in \emph{Proceedings of the IEEE
  Conference on Computer Vision and Pattern Recognition (CVPR)}, 2016, pp.
  2536--2544.

\bibitem{iizuka2017globally}
S.~Iizuka, E.~Simo-Serra, and H.~Ishikawa, ``Globally and locally consistent
  image completion,'' \emph{ACM Transactions on Graphics (ToG)}, vol.~36,
  no.~4, pp. 1--14, 2017.

\bibitem{yu2018generative}
J.~Yu, Z.~Lin, J.~Yang, X.~Shen, X.~Lu, and T.~S. Huang, ``Generative image
  inpainting with contextual attention,'' in \emph{Proceedings of the IEEE
  Conference on Computer Vision and Pattern Recognition (CVPR)}, 2018, pp.
  5505--5514.

\bibitem{zheng2019pluralistic}
C.~Zheng, T.-J. Cham, and J.~Cai, ``Pluralistic image completion,'' in
  \emph{Proceedings of the IEEE Conference on Computer Vision and Pattern
  Recognition (CVPR)}, 2019, pp. 1438--1447.

\bibitem{yan2019PENnet}
Y.~Zeng, J.~Fu, H.~Chao, and B.~Guo, ``Learning pyramid-context encoder network
  for high-quality image inpainting,'' in \emph{The IEEE Conference on Computer
  Vision and Pattern Recognition (CVPR)}, June 2019.

\bibitem{li2017generative}
Y.~Li, S.~Liu, J.~Yang, and M.-H. Yang, ``Generative face completion,'' in
  \emph{Proceedings of the IEEE Conference on Computer Vision and Pattern
  Recognition (CVPR)}, 2017, pp. 3911--3919.

\bibitem{liao2018face}
H.~Liao, G.~Funka-Lea, Y.~Zheng, J.~Luo, and S.~K. Zhou, ``Face completion with
  semantic knowledge and collaborative adversarial learning,'' in \emph{Asian
  Conference on Computer Vision}.\hskip 1em plus 0.5em minus 0.4em\relax
  Springer, 2018, pp. 382--397.

\bibitem{isola2017image}
P.~Isola, J.-Y. Zhu, T.~Zhou, and A.~A. Efros, ``Image-to-image translation
  with conditional adversarial networks,'' in \emph{Proceedings of the IEEE
  Conference on Computer Vision and Pattern Recognition (CVPR)}, 2017, pp.
  1125--1134.

\bibitem{liu2015faceattributes}
Z.~Liu, P.~Luo, X.~Wang, and X.~Tang, ``Deep learning face attributes in the
  wild,'' in \emph{Proceedings of International Conference on Computer Vision
  (ICCV)}, December 2015.

\bibitem{karras2018progressive}
\BIBentryALTinterwordspacing
T.~Karras, T.~Aila, S.~Laine, and J.~Lehtinen, ``Progressive growing of {GAN}s
  for improved quality, stability, and variation,'' in \emph{International
  Conference on Learning Representations (ICLR)}, 2018. [Online]. Available:
  \url{https://openreview.net/forum?id=Hk99zCeAb}
\BIBentrySTDinterwordspacing

\bibitem{bertalmio2000image}
M.~Bertalmio, G.~Sapiro, V.~Caselles, and C.~Ballester, ``Image inpainting,''
  in \emph{Proceedings of the 27th annual conference on Computer graphics and
  interactive techniques}.\hskip 1em plus 0.5em minus 0.4em\relax ACM
  Press/Addison-Wesley Publishing Co., 2000, pp. 417--424.

\bibitem{ballester2001filling}
C.~Ballester, M.~Bertalmio, V.~Caselles, G.~Sapiro, and J.~Verdera,
  ``Filling-in by joint interpolation of vector fields and gray levels,''
  \emph{IEEE transactions on image processing}, vol.~10, no.~8, pp. 1200--1211,
  2001.

\bibitem{bertalmio2003simultaneous}
M.~Bertalmio, L.~Vese, G.~Sapiro, and S.~Osher, ``Simultaneous structure and
  texture image inpainting,'' \emph{IEEE transactions on image processing},
  vol.~12, no.~8, pp. 882--889, 2003.

\bibitem{jin2015annihilating}
K.~H. Jin and J.~C. Ye, ``Annihilating filter-based low-rank hankel matrix
  approach for image inpainting,'' \emph{IEEE Transactions on Image
  Processing}, vol.~24, no.~11, pp. 3498--3511, 2015.

\bibitem{dobrosotskaya2008wavelet}
J.~A. Dobrosotskaya and A.~L. Bertozzi, ``A wavelet-laplace variational
  technique for image deconvolution and inpainting,'' \emph{IEEE Transactions
  on Image Processing}, vol.~17, no.~5, pp. 657--663, 2008.

\bibitem{qin2013novel}
C.~Qin, C.-C. Chang, and Y.-P. Chiu, ``A novel joint data-hiding and
  compression scheme based on smvq and image inpainting,'' \emph{IEEE
  transactions on image processing}, vol.~23, no.~3, pp. 969--978, 2013.

\bibitem{xu2010image}
Z.~Xu and J.~Sun, ``Image inpainting by patch propagation using patch
  sparsity,'' \emph{IEEE transactions on image processing}, vol.~19, no.~5, pp.
  1153--1165, 2010.

\bibitem{criminisi2004region}
A.~Criminisi, P.~P{\'e}rez, and K.~Toyama, ``Region filling and object removal
  by exemplar-based image inpainting,'' \emph{IEEE Transactions on Image
  Processing}, vol.~13, no.~9, pp. 1200--1212, 2004.

\bibitem{li2014color}
Z.~Li, H.~He, H.-M. Tai, Z.~Yin, and F.~Chen, ``Color-direction
  patch-sparsity-based image inpainting using multidirection features,''
  \emph{IEEE Transactions on Image Processing}, vol.~24, no.~3, pp. 1138--1152,
  2014.

\bibitem{liu2012exemplar}
Y.~Liu and V.~Caselles, ``Exemplar-based image inpainting using multiscale
  graph cuts,'' \emph{IEEE transactions on image processing}, vol.~22, no.~5,
  pp. 1699--1711, 2012.

\bibitem{li2014universal}
F.~Li and T.~Zeng, ``A universal variational framework for sparsity-based image
  inpainting,'' \emph{IEEE Transactions on Image Processing}, vol.~23, no.~10,
  pp. 4242--4254, 2014.

\bibitem{ballester2007inpainting}
C.~Ballester, M.~Bertalm{\'\i}o, V.~Caselles, L.~Garrido, A.~Marques, and
  F.~Ranchin, ``An inpainting-based deinterlacing method,'' \emph{IEEE
  Transactions on Image Processing}, vol.~16, no.~10, pp. 2476--2491, 2007.

\bibitem{barnes2009patchmatch}
C.~Barnes, E.~Shechtman, A.~Finkelstein, and D.~B. Goldman, ``Patchmatch: A
  randomized correspondence algorithm for structural image editing,'' in
  \emph{ACM Transactions on Graphics (ToG)}, vol.~28.\hskip 1em plus 0.5em
  minus 0.4em\relax ACM, 2009, p.~24.

\bibitem{whyte2009get}
O.~Whyte, J.~Sivic, and A.~Zisserman, ``Get out of my picture! internet-based
  inpainting.'' in \emph{British Machine Vision Conference (BMVC)}, vol.~2,
  2009, p.~5.

\bibitem{he2016deep}
K.~He, X.~Zhang, S.~Ren, and J.~Sun, ``Deep residual learning for image
  recognition,'' in \emph{Proceedings of the IEEE Conference on Computer Vision
  and Pattern Recognition (CVPR)}, 2016, pp. 770--778.

\bibitem{yang2017high}
C.~Yang, X.~Lu, Z.~Lin, E.~Shechtman, O.~Wang, and H.~Li, ``High-resolution
  image inpainting using multi-scale neural patch synthesis,'' in
  \emph{Proceedings of the IEEE Conference on Computer Vision and Pattern
  Recognition}, 2017, pp. 6721--6729.

\bibitem{liu2018image}
G.~Liu, F.~A. Reda, K.~J. Shih, T.-C. Wang, A.~Tao, and B.~Catanzaro, ``Image
  inpainting for irregular holes using partial convolutions,'' in
  \emph{Proceedings of the European Conference on Computer Vision (ECCV)},
  2018, pp. 85--100.

\bibitem{yan2018shift}
Z.~Yan, X.~Li, M.~Li, W.~Zuo, and S.~Shan, ``Shift-net: Image inpainting via
  deep feature rearrangement,'' in \emph{Proceedings of the European Conference
  on Computer Vision (ECCV)}, 2018, pp. 1--17.

\bibitem{ren2019structureflow}
Y.~Ren, X.~Yu, R.~Zhang, T.~H. Li, S.~Liu, and G.~Li, ``Structureflow: Image
  inpainting via structure-aware appearance flow,'' in \emph{Proceedings of the
  IEEE International Conference on Computer Vision}, 2019, pp. 181--190.

\bibitem{hong2019deep}
X.~Hong, P.~Xiong, R.~Ji, and H.~Fan, ``Deep fusion network for image
  completion,'' in \emph{Proceedings of the 27th ACM International Conference
  on Multimedia}, 2019, pp. 2033--2042.

\bibitem{xie2019image}
C.~Xie, S.~Liu, C.~Li, M.-M. Cheng, W.~Zuo, X.~Liu, S.~Wen, and E.~Ding,
  ``Image inpainting with learnable bidirectional attention maps,'' in
  \emph{Proceedings of the IEEE International Conference on Computer Vision},
  2019, pp. 8858--8867.

\bibitem{xiong2019foreground}
W.~Xiong, J.~Yu, Z.~Lin, J.~Yang, X.~Lu, C.~Barnes, and J.~Luo,
  ``Foreground-aware image inpainting,'' in \emph{Proceedings of the IEEE
  Conference on Computer Vision and Pattern Recognition}, 2019, pp. 5840--5848.

\bibitem{song2018contextual}
Y.~Song, C.~Yang, Z.~Lin, X.~Liu, Q.~Huang, H.~Li, and C.-C. Jay~Kuo,
  ``Contextual-based image inpainting: Infer, match, and translate,'' in
  \emph{Proceedings of the European Conference on Computer Vision (ECCV)},
  2018, pp. 3--19.

\bibitem{wang2018image}
Y.~Wang, X.~Tao, X.~Qi, X.~Shen, and J.~Jia, ``Image inpainting via generative
  multi-column convolutional neural networks,'' in \emph{Advances in Neural
  Information Processing Systems}, 2018, pp. 331--340.

\bibitem{yu2019free}
J.~Yu, Z.~Lin, J.~Yang, X.~Shen, X.~Lu, and T.~S. Huang, ``Free-form image
  inpainting with gated convolution,'' in \emph{Proceedings of the IEEE
  International Conference on Computer Vision}, 2019, pp. 4471--4480.

\bibitem{ding2018image}
D.~Ding, S.~Ram, and J.~J. Rodr{\'\i}guez, ``Image inpainting using nonlocal
  texture matching and nonlinear filtering,'' \emph{IEEE Transactions On Image
  Processing}, vol.~28, no.~4, pp. 1705--1719, 2018.

\bibitem{xue2017depth}
H.~Xue, S.~Zhang, and D.~Cai, ``Depth image inpainting: Improving low rank
  matrix completion with low gradient regularization,'' \emph{IEEE Transactions
  on Image Processing}, vol.~26, no.~9, pp. 4311--4320, 2017.

\bibitem{buyssens2016depth}
P.~Buyssens, O.~Le~Meur, M.~Daisy, D.~Tschumperl{\'e}, and O.~L{\'e}zoray,
  ``Depth-guided disocclusion inpainting of synthesized rgb-d images,''
  \emph{IEEE Transactions on Image Processing}, vol.~26, no.~2, pp. 525--538,
  2016.

\bibitem{yeh2017semantic}
R.~A. Yeh, C.~Chen, T.~Yian~Lim, A.~G. Schwing, M.~Hasegawa-Johnson, and M.~N.
  Do, ``Semantic image inpainting with deep generative models,'' in
  \emph{Proceedings of the IEEE Conference on Computer Vision and Pattern
  Recognition (CVPR)}, 2017, pp. 5485--5493.

\bibitem{vitoria2018semantic}
P.~Vitoria, J.~Sintes, and C.~Ballester, ``Semantic image inpainting through
  improved wasserstein generative adversarial networks,'' \emph{arXiv preprint
  arXiv:1812.01071}, 2018.

\bibitem{zhang2018semantic}
H.~Zhang, Z.~Hu, C.~Luo, W.~Zuo, and M.~Wang, ``Semantic image inpainting with
  progressive generative networks,'' in \emph{ACM Multimedia Conference on
  Multimedia Conference (ACM MM)}.\hskip 1em plus 0.5em minus 0.4em\relax ACM,
  2018, pp. 1939--1947.

\bibitem{xiao2019cisi}
J.~Xiao, L.~16~Liao, Q.~Liu, and R.~Hu, ``Cisi-net: Explicit latent content
  inference and imitated style rendering for image inpainting,'' in
  \emph{Proceedings of the AAAI Conference on Artificial Intelligence (AAAI)},
  vol.~33, 2019, pp. 354--362.

\bibitem{doersch2016tutorial}
C.~DOERSCH, ``Tutorial on variational autoencoders,'' \emph{stat}, vol. 1050,
  p.~13, 2016.

\bibitem{pu2016variational}
Y.~Pu, Z.~Gan, R.~Henao, X.~Yuan, C.~Li, A.~Stevens, and L.~Carin,
  ``Variational autoencoder for deep learning of images, labels and captions,''
  in \emph{Advances in neural information processing systems}, 2016, pp.
  2352--2360.

\bibitem{lutz2018alphagan}
S.~Lutz, K.~Amplianitis, and A.~Smolic, ``Alphagan: Generative adversarial
  networks for natural image matting.''

\bibitem{makhzani2015adversarial}
A.~Makhzani, J.~Shlens, N.~Jaitly, I.~Goodfellow, and B.~Frey, ``Adversarial
  autoencoders.''

\bibitem{rosca2017variational}
M.~Rosca, B.~Lakshminarayanan, D.~Warde-Farley, and S.~Mohamed, ``Variational
  approaches for auto-encoding generative adversarial networks.''

\bibitem{wang2018facial}
N.~Wang, X.~Gao, D.~Tao, H.~Yang, and X.~Li, ``Facial feature point detection:
  A comprehensive survey,'' \emph{Neurocomputing}, vol. 275, pp. 50--65, 2018.

\bibitem{baidu2019}
Baidu, \url{https://ai.baidu.com/tech/body/seg}, 2019.

\bibitem{kingma2015adam}
D.~Kingma and J.~Ba, ``Adam: A method for stochastic optimization in:
  Proceedings of international conference on learning representations,'' 2015.

\end{thebibliography}

%\clearpage
\appendices
\section{DEGNet Architecture}

DEGNet includes three key components: domain embedding network, domain embedded generator, and discriminator. More specifically, a domain embedding network consists of two encoders and three decoders. We use FCN as the domain embedded generator. The discriminators include the patch discriminator and the global discriminator. The detailed structure of DEGNet is shown in Table~\ref{tab1}.

\begin{table}[t]
\centering
\caption{Detailed structure of DEGNet.}\label{tab1}
\scalebox{0.8} {
\begin{tabular}{lcccccc}
\hline
\textbf{Operation} & \textbf{Kernel} & \textbf{Strides} & \textbf{Filters} & \textbf{BN} & \textbf{Activation} & \textbf{Output to}\\
\hline
\multicolumn{7}{l}{\textbf{Encoder--Input: $N_{batch}\times256\times256\times3$}}\\
\hline
1-a Conv2D & 4*4 & 2*2 & 32 & No & LeakyReLu & 1-b \\
1-b Conv2D & 4*4 & 2*2 & 64 & No & LeakyReLu & 1-c \\
1-c Conv2D & 4*4 & 2*2 & 128 & No & LeakyReLu & 1-d\\
1-d Conv2D & 4*4 & 2*2 & 128 & No & LeakyReLu & 1-e\\
1-e Conv2D & 4*4 & 2*2 & 128 & No & LeakyReLu & 1-f,1-g\\
1-f Dense & $--$ & $--$ & 256 & No & Sigmoid & $--$\\
1-g Dense & $--$ & $--$ & 256 & No & Sigmoid & $--$\\
\hline
\multicolumn{7}{l}{\textbf{Decoder--Input: $N_{batch}\times512$}}\\
\hline
2-a Reshape &$--$ &$--$ &$--$ &$--$ &$--$ &(16,16,2) 2-b\\
2-b Deconv2D & 4*4 & 2*2 & 128 & No & ReLu & 2-c \\
2-c Deconv2D & 4*4 & 2*2 & 64 & No & ReLu & 2-d\\
2-d Deconv2D & 4*4 & 2*2 & 32 & No & ReLu & 2-e\\
2-e Deconv2D & 4*4 & 2*2 & 16 & No & ReLu & 2-f\\
2-f Conv2D & 1*1 & 1*1 & 3 & No & Sigmoid & $--$\\

\hline
\multicolumn{7}{l}{\textbf{FCN--Input: $N_{batch}\times256\times256\times3$(3-b), $N_{batch}\times2512$(3-a)}}\\
\hline
3-a Reshape  &$--$	&$--$   &$--$	&$--$	 &$--$       &(16,16,2)3-L\\
3-b Conv2D	 &3*3	&1*1	&32	    &No	 &ReLu	   &3-c\\
3-c Conv2D	 &3*3	&1*1	&32	    &No	 &ReLu	   &3-d,3-W\\
3-d Conv2D	 &3*3	&2*2	&64	    &No	 &ReLu	   &3-e\\
3-e Conv2D	 &3*3	&1*1	&64	    &No	 &ReLu	   &3-f,3-T\\
3-f Conv2D	 &3*3	&2*2	&128	&No	 &ReLu	   &3-g\\
3-g Conv2D	 &3*3	&1*1	&128	&No	 &ReLu	   &3-h,3-Q\\
3-h Conv2D	 &3*3	&2*2	&256	&No	 &ReLu	   &3-I\\
3-i Conv2D	 &3*3	&1*1	&256	&No	 &ReLu	   &3-j,3-N\\
3-j Conv2D	 &3*3	&2*2	&512	&No	 &ReLu	   &3-k\\
3-k Conv2D	 &3*3	&1*1	&512	&No	 &ReLu	   &3-L\\
3-L Concat	 &$--$	&$--$	&$--$	&$--$&$--$	   &3-M\\
3-M Deconv2D &3*3   &2*2    &256	&No	 &ReLu     &3-N\\
3-N Concat	 &$--$  &$--$	&$--$	&$--$&$--$	   &3-O\\
3-O Conv2D	 &3*3   &1*1    &256    &No  &ReLu     &3-P\\
3-P DeConv2D &3*3   &2*2    &128	&No	 &ReLu     &3-Q\\
3-Q Concat	 &$--$	&$--$   &$--$   &$--$ &$--$	   &3-R\\
3-R Conv2D	 &3*3   &1*1    &128    &No  &ReLu     &3-S\\
3-S DeConv2D &3*3   &2*2    &64	    &No	 &ReLu     &3-T\\
3-T Concat	 &$--$	&$--$	&$--$	&$--$&$--$	   &3-U\\
3-U Conv2D	 &3*3	&1*1	&64	    &No	 &ReLu     &3-V\\
3-V Deconv2D &3*3	&2*2	&32	    &No	 &ReLu     &3-W\\
3-W Concat	 &$--$	&$--$   &$--$   &$--$&$--$	   &3-X\\
3-X Conv2D	&3*3	&1*1	&32	    &No	 &ReLu	   &3-Y\\
3-Y Conv2D	&3*3	&1*1	&3	    &No	 &Sigmoid  &$--$\\

\hline
\multicolumn{7}{l}{\textbf{Dp--Input: $N_{batch}\times256\times256\times3$}}\\
\hline
4-a Conv2D	&4*4	&2*2	&32	  &No	&LeakyReLu	& 4-b\\
4-b Conv2D	&4*4	&2*2	&64	  &No	&LeakyReLu	&4-c\\
4-c Conv2D	&4*4	&2*2	&128  &No	&LeakyReLu	&4-d\\
4-d Conv2D	&4*4	&2*2	&256  &No	&LeakyReLu	&4-e\\
4-e Conv2D	&4*4	&2*2	&512  &No	&LeakyReLu	&4-f\\
4-f Conv2D	&1*1	&1*1	&1	  &No	&sigmoid	&--\\

\hline
\multicolumn{7}{l}{\textbf{Dg--Input: $N_{batch}\times256\times256\times3$}}\\
\hline
5-a Conv2D	&4*4	&2*2	&32	  &No	&LeakyReLu	&5-b\\
5-b Conv2D	&4*4	&2*2	&64	  &No	&LeakyReLu	&5-c\\
5-c Conv2D	&4*4	&2*2	&128  &No	&LeakyReLu	&5-d\\
5-d Conv2D	&4*4	&2*2	&256  &No	&LeakyReLu	&5-e\\
5-e Conv2D	&4*4	&2*2	&512  &No	&LeakyReLu	&5-f\\
5-f Dense	&$--$   &$--$	&1	  &$--$	&sigmoid   &$--$\\
\hline
\end{tabular}
}
\end{table}
\end{document}